\pdfoutput=1

\documentclass[11pt]{article}

\usepackage{EMNLP2023}

\usepackage{times}
\usepackage{booktabs}
\usepackage{latexsym}
\usepackage{graphicx}
\usepackage{amsmath}
\usepackage{multirow}
\usepackage{xcolor}

\usepackage[T1]{fontenc}
\usepackage[utf8]{inputenc}
\usepackage{microtype}
\usepackage{inconsolata}

\graphicspath{{}}

\newcommand{\new}[1]{{\color{black}#1}}
\newcommand{\rev}[1]{{\color{black}#1}}

\newcommand{\benchmark}{CARTE}
\newcommand{\Lbench}{CARTE-LV}
\newcommand{\numberquestions}{3,132}
\newcommand{\numbergenerated}{4,354}
\newcommand{\numberfiltered}{3,173}
\newcommand{\numbermodels}{32}

\newcommand{\nLV}{220}

\newcommand{\nSuspicious}{190}

\newcommand{\bestModel}{Luciole-23B-Instruct}
\newcommand{\bestAccZero}{75.6}
\newcommand{\bestAccThree}{75.9}

\title{CARTE: A Benchmark for Mapping Language Model\\ Knowledge Across France}

\author{
Sarah Almeida Carneiro$^{1}$,
Christos Xypolopoulos$^{1,2}$, Xiao Fei$^{1}$, \\
\textbf{Yang Zhang$^{1}$, Michalis Vazirgiannis$^{1,3}$} \\
$^{1}$École Polytechnique, Institut Polytechnique de Paris, France \\
$^{2}$National Technical University of Athens,\\
$^{3}$Mohamed bin Zayed University of Artificial Intelligence, United Arab Emirates \\
\small
\texttt{\{sarah.almeida-carneiro, christos.xypolopoulos, michalis.vazirgiannis\}@polytechnique.edu}
}

\begin{document}
\maketitle
\begin{abstract}
We introduce \benchmark\footnote{\url{https://huggingface.co/datasets/ScarAlcar/CARTE}} (\textbf{C}ulturally \textbf{A}nchored \textbf{R}egional-\textbf{T}erritorial \textbf{E}valuation), a multiple-choice benchmark for evaluating the ability of large language models (LLMs) to \new{discriminate fine-grained, geographically grounded and regionally differentiated knowledge} within France. While prior benchmarks focus on national-level cultural understanding, they largely overlook intra-country variation and the need to distinguish between closely related regional contexts. \benchmark\ addresses this gap by introducing \new{\numberquestions} questions spanning the 13 metropolitan regions of France and covering \new{12} thematic domains, including culture, language, demographics, economy, environment, and mobility. \new{Questions underwent four-stage validation: human review, LLM-as-a-judge, model-based difficulty calibration, and a second human review.} We further introduce \Lbench, a subset targeting \textbf{L}inguistic \textbf{V}ariation across French regions, enabling focused evaluation of language-related differences. \new{We evaluate \numbermodels\ open-weight LLMs of 1B to 30B parameters and three frontier API models across zero-, one-, and three-shot settings.} Our experiments reveal performance disparities across regions and model scales, suggesting systematic gaps in pretraining coverage and limited robustness to intra-national variation.

\end{abstract}

\section{Introduction}

\begin{figure}[t!]
  \centering
  \includegraphics[width=\columnwidth]{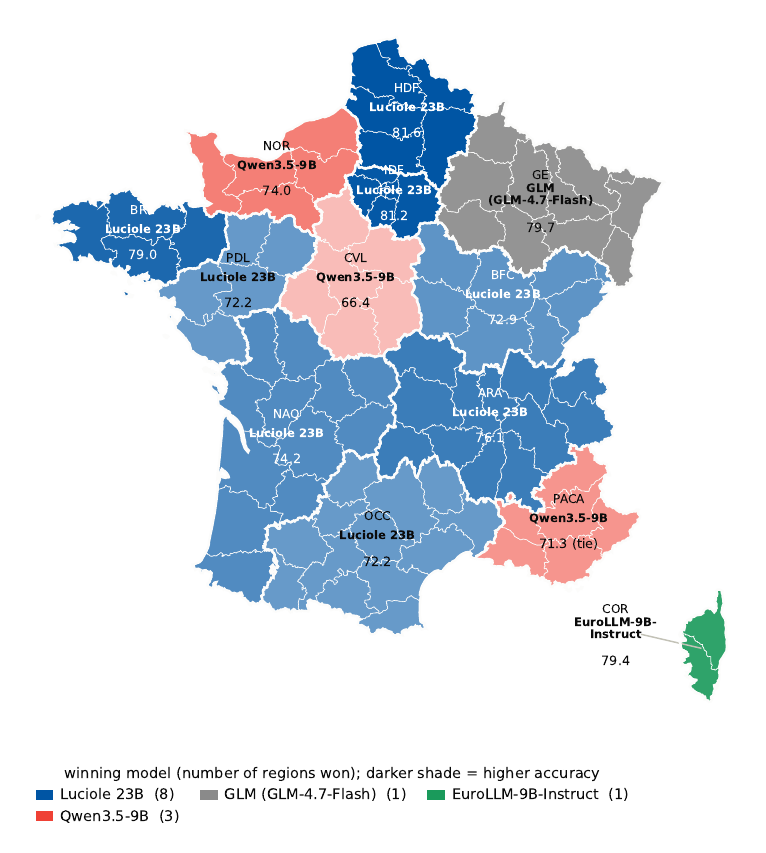}
 \caption{\new{Best model per region at 0-shot on \benchmark, over the 32 open-weight models run locally: each region is colored by the model with the highest accuracy on its questions and labeled with that model and its accuracy; a darker shade means a higher accuracy.}}
  \label{fig:Best_pre_region}
\end{figure}

Recent years have seen rapid growth in the development of sovereign large language models (LLMs) across multiple countries. However, the evaluation of these models still relies predominantly on English-language benchmarks or datasets translated from English into the target language~\cite{thellmann2024towards, min2025towards, guo2025large}. Although modern translation systems preserve contextual meaning more effectively than previous approaches, culturally grounded concepts such as traditions, idiomatic expressions, gastronomy, and regional references often remain difficult to translate faithfully~\cite{al2025bridging, naveen2024overview, myung2024blend}.

This limitation particularly affects French LLM evaluation. Despite French not being considered a low-resource language, most existing benchmarks rely on translated general-purpose datasets such as MMLU~\cite{hendrycks2021mmlu, ying2025disentangling}. Although such benchmarks remain effective for evaluating domains with relatively universal knowledge representations, including mathematics, physics, biology, or general facts, they provide limited insight into culturally localized and everyday knowledge. As a result, current evaluation frameworks only partially assess whether models capture the social, historical, and regional specificities of French culture.

Recent work on culturally grounded evaluation for lower-resource languages~\cite{zhang2026greekmmlu, koto2024arabicmmlu, yuksel2024turkishmmlu}, has highlighted the importance of localized benchmarks for measuring cultural alignment beyond general reasoning capabilities. Motivated by these efforts, and by the global importance of French as one of the six official languages of the United Nations, we introduce \benchmark\ a \textbf{C}ulturally \textbf{A}nchored \textbf{R}egional-\textbf{T}erritorial \textbf{E}valuation benchmark designed specifically for France.

\benchmark\ is fully French and comprises \new{\numberquestions} multiple-choice questions covering \new{12} broad thematic domains, including culture, economy, environment, language, and mobility. Rather than testing general facts, the objective of this benchmark is to evaluate fine-grained, region-specific knowledge and intra-national variations across the 13 metropolitan regions of France, nuances that are exceptionally difficult to assess through translated or English-centric benchmarks alone. To ensure dataset quality and territorial relevance, we utilized a multi-stage validation pipeline combining \new{human verification by volunteers, LLM-based filtering guided by the human annotations, and a model-based difficulty calibration so that the most suspicious items were sent back to the reviewers}.
We then evaluate \new{\numbermodels} open-weight general-purpose, European-developed, and French-focused LLMs \new{of 1B to 30B parameters, together with three frontier API models (Section~\ref{sec:models}),} using multiple-choice scoring to analyze their geographic alignment and ability to reason about closely related territorial contexts.

Our main contributions are summarized as follows.

\begin{itemize}
    \item We introduce \textbf{\benchmark}, covering \new{12} distinct dimensions of cultural knowledge. This suite maps all metropolitan regions of France, moving beyond uniform national stereotypes to capture true regional diversity.

    \item We additionally introduce \textbf{\Lbench}, a subset of \benchmark, that can be used as a standalone benchmark focused exclusively on regional linguistic variation inside France, enabling fine-grained linguistic evaluation across regions.

    \item \new{We describe a \textbf{validation pipeline} in which human volunteers annotate 25\% of the generated questions, their verdicts and comments calibrate an LLM-as-a-judge filter over the remaining items, and the questions that more than 90\% of the evaluated models fail are re-validated by humans.}

    \item We benchmark \textbf{\new{\numbermodels}\ open-weight LLMs} and three frontier API models, systematically contrasting targeted French-native and European open-weight models against state-of-the-art, English-centric frontier models. We further analyze their cultural reasoning robustness across zero-, one-, and three-shot in-context learning paradigms.

    \item We present a \textbf{specialized evaluation of regional linguistic nuances} via \Lbench. 
\end{itemize}

These contributions expose limitations in current evaluation practices, where translated benchmarks often do not reveal gaps in culturally grounded knowledge. By providing a structured framework for French cultural evaluation, this work enables a more reliable assessment of how well models generalize beyond English-centric data distributions. Ultimately, this supports the development of LLMs that are more sensitive to cultural context and better aligned with the linguistic and societal realities of their target users.

\section{Related Work}

\paragraph{General Language Models:} Recent progress in LLMs has been driven by decoder-only transformer architectures trained on large-scale corpora. Models such as Qwen3~\cite{yang2025qwen3}, Llama~\cite{grattafiori2024llama3}, Gemma~\cite{gemmateam2026gemma4}, BLOOM~\cite{workshop2022bloom}, \new{GLM~\cite{glm2024chatglm}} and Gemini~\cite{team2023gemini} have demonstrated strong multilingual and reasoning capabilities. Their performance is largely enabled by training on massive and diverse internet-scale datasets, allowing them to generalize across a broad range of tasks including question answering, summarization, translation, and code generation.

General-purpose LLMs are particularly effective in English-language settings, as a significant proportion of publicly available high-quality web data, academic publications, technical documentation, and digital resources are predominantly written in English~\cite{nguyen2024democratizing, li2024culturellm}. Although multilingual coverage has improved substantially with increasingly diverse web crawling and multilingual datasets, the distribution and quality of training data remain uneven across languages and regions.

As a consequence, while these models exhibit strong general reasoning abilities, broad knowledge does not necessarily translate into deep expertise in all. In highly specialized, low-resource, or region-specific contexts, especially those involving nuanced cultural, linguistic, legal, or technical knowledge, general-purpose models may exhibit weaker performance when compared to domain-adapted or fine-tuned systems~\cite{joshi2024fine}. This limitation becomes more evident for niche subdomains or highly localized targets where relevant training data is scarce or underrepresented.

\paragraph{French Language Models:}

Mistral~\cite{jiang2023mistral} demonstrated the competitiveness of efficient open-weight decoder-only architectures and became a foundation for many French fine-tuned assistants. CroissantLLM~\cite{faysse2025croissantllm} focused on multilingual and French-centric pretraining with open data pipelines, while Lucie~\cite{gouvert2025lucie} and the Luciole family~\cite{openllmfrance2026luciole} emphasized transparent and fully open French language model development. Claire~\cite{louradour2024claire} introduced conversationally oriented French language modeling resources targeting dialogue applications. More recent instruction-tuned French assistants such as Luth~\cite{lasbordes2026luth} and Vigogne~\cite{vigogne} further explored alignment, conversational fine-tuning, and French-specific assistant behavior. \new{The Gaperon suite~\cite{godey2025gaperon} released bilingual English--French base and instruction-tuned models.}

These efforts highlight the growing interest in developing language technologies tailored to French linguistic and cultural contexts. However, despite being trained primarily on French corpora or developed by French-speaking communities, such models may still inherit important limitations related to data imbalance and uneven regional representation~\cite{xu2025survey}. As a result, models optimized for French do not necessarily guarantee strong performance across all Francophone contexts. Subtle deficiencies may emerge when dealing with highly localized information, region-specific terminology, or underrepresented communities. Identifying these weaknesses is further complicated by the current lack of comprehensive regional and culturally grounded evaluation benchmarks for French and broader Francophone NLP. Without sufficiently granular benchmarks, it becomes difficult for researchers to determine which linguistic varieties, regions, domains, or cultural contexts remain underrepresented during training and evaluation, limiting the ability to systematically improve coverage and robustness.

\paragraph{LLM Evaluation:}
Existing benchmarks such as FQuAD~\cite{dhoffschmidt-etal-2020-fquad}, mMARCO~\cite{mMARCO2021}, Belebele~\cite{bandarkar-etal-2023-belebele}, and MMLU~\cite{hendrycks2021mmlu} assess general language understanding, reasoning, reading comprehension, and academic knowledge, but do not specifically target regional cultural, social, or administrative knowledge. French conversational resources such as CFDD~\cite{hunter2023claire} and COLE~\cite{beauchemin2025cole} evaluate dialogue fluency and NLU, respectively, with limited emphasis on regional variation and cultural references.

Recent benchmarks explore cultural awareness in LLMs~\cite{chiu2024culturalbencharxiv, myung2024blend, romanou2025include}, but typically focus on national or language-level differences. Sub-national approaches include BertaQA~\cite{etxaniz2024bertaqaneurips}, which compares Basque-local and global knowledge; IndoCulture~\cite{koto2024indoculture}, covering eleven Indonesian provinces; and SaudiCulture~\cite{ayash2025saudiculturejksu}, which examines regional knowledge in Saudi Arabia. \rev{\benchmark\ builds on this line of work while offering three distinctions: coverage of all 13 metropolitan French administrative regions under a shared 12-topic taxonomy, enabling direct regional comparisons; a dedicated subset for linguistic variation; and a validated resource in which human annotations calibrate automatic filtering and model failures trigger a second human review. Thus, our contribution is not cultural evaluation itself, but its regional resolution and the validation of the resulting benchmark.} To our knowledge, \benchmark\ is the first benchmark to provide comprehensive evaluation of LLM knowledge and cultural understanding across all 13 metropolitan regions of France, enabling systematic analysis of regional performance disparities.

\section{Benchmark Curation Methodology}

In the following subsections, we introduce the development of a regionally grounded benchmark specifically designed for France. The benchmark consists of multiple-choice questions with validated answers spanning \new{12} thematic domains. Its primary objective is to move beyond translated or English-centric evaluation paradigms by providing a culturally and territorially aligned framework for assessing French language models. Additionally, we introduce a complementary regional-language subset of \benchmark, namely \Lbench, which can serve as a standalone evaluation resource.

Overall, \benchmark\ is designed to assess whether models exhibit genuine alignment with French socio-cultural and geographic contexts, and to enable a finer-grained analysis of knowledge across France’s internal regional divisions.

\subsection{Data Collection and Curation}

\benchmark\ was constructed from a corpus of manually selected, human-authored French documents gathered from open-access institutional repositories and distributed under non-commercial creative and research licenses. The selection process was designed to ensure diversity across both topical domains and regional variants of French.
The corpus spans the following thematic areas: institutions and governance; economy and industry; agriculture and terroirs; infrastructure and networks; transport and mobility; environment and biodiversity; territory and spatial planning; society and social realities; demography; language and linguistics; history and heritage; and culture, traditions, and society. For further details on what is covered in these major topics see Appendix~\ref{ap:TopicsDetail}.

In addition, \benchmark\ was curated to reflect the diversity of France’s 13 metropolitan administrative regions: Auvergne-Rhône-Alpes, Bourgogne-Franche-Comté, Bretagne, Centre-Val de Loire, Corse, Grand Est, Hauts-de-France, Île-de-France, Normandie, Nouvelle-Aquitaine, Occitanie, Pays de la Loire, and Provence-Alpes-Côte d’Azur. \new{Questions whose scope is the whole country (national institutions, nationwide practices) carry the region tag \textit{France}.}

\subsection{Question Generation}

To ensure the generation of up-to-date and contextually grounded evaluation items, we did not rely on existing internet-sourced quiz questions. Although such questions are typically well-formed and factually reliable, they predominantly cover widely known and readily accessible information that is already strongly represented in large-scale language models. As a result, they are often outdated with respect to evolving information and insufficiently capture fine-grained, region-specific content coverage.

Instead, we constructed a corpus of \new{\numbergenerated\ candidate} multiple-choice questions (MCQs) using a document-grounded augmentation pipeline based on Gemini 3 Flash. Source documents were first manually reviewed to ensure linguistic quality, factual consistency, and regional relevance. Questions were then generated conditionally on these validated documents to promote diversity, reduce reliance on memorized knowledge, and better capture localized and region-specific information.

The prompting protocol enforced several constraints (for more detail on the prompt see Appendix~\ref{ap:MCQprompt}):
\begin{enumerate}
    \item The model is assigned an expert role covering French territorial, socio-economic, cultural, linguistic, and environmental domains, and generates questions grounded exclusively in a provided document.

    \item The task consists of producing autonomous multiple-choice questions (QCM) on France or the French language, each requiring explicit territorial anchoring and, when applicable, precise temporal references.

    \item Questions must be reformulated as general knowledge items, independent of document structure, and designed to be analytical, non-trivial, and diverse across themes, scales, and cognitive levels.

    \item Each question includes five answer choices (A--E): one correct answer, three plausible distractors and one abstention option, always ``I don't know'' (\textit{``Je ne sais pas''}); distractors must be realistic and geographically coherent, with only one correct answer.

    \item Strong variation is enforced across question types, spatial scales, and thematic domains to avoid repetition.
\end{enumerate}

Correct answers are uniformly distributed across positions A--E, and both answer and abstention positions are shuffled to prevent positional bias. Distractors are realistic, often drawn from other French or European regions, yet distinct enough to ensure a single unambiguous answer. Questions also vary in format (e.g., comparison, identification, and spatial reasoning), geographic scale, and topic to limit structural bias and repetition; Table~\ref{tab:questions_region_category} (Appendix~\ref{ap:qpc}) reports the final distribution across regions and categories.

\begin{figure*}[t]
  \centering
  \includegraphics[width=\textwidth]{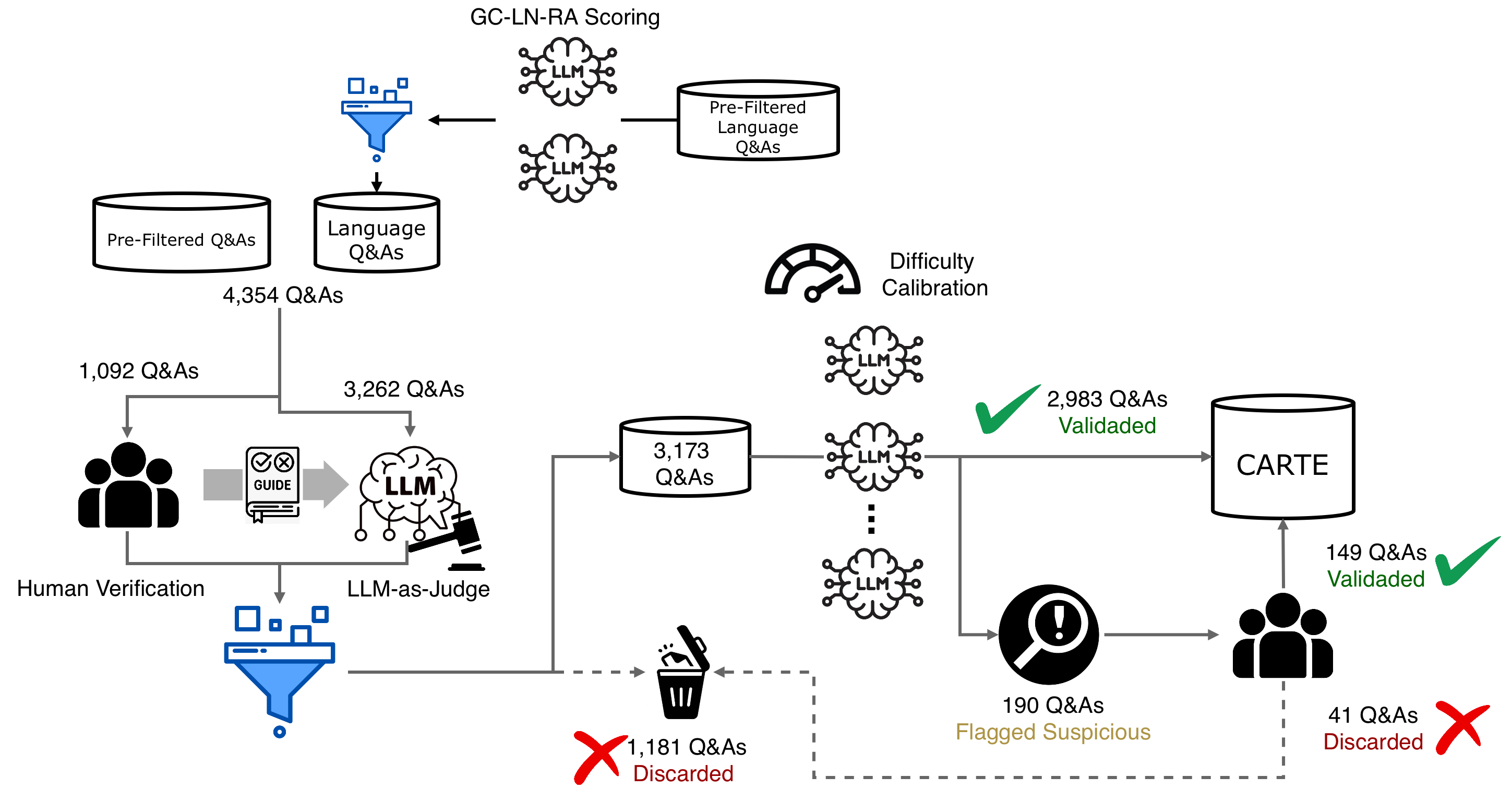}
  \caption{\new{Construction and validation pipeline of \benchmark: number of questions after each stage, from the \numbergenerated\ generated items to the \numberquestions\ questions}}
  \label{fig:pipeline}
\end{figure*}

\subsection[Human Validation]{\new{Human Validation}}
\label{sec:human-validation}

\new{Figure~\ref{fig:pipeline} summarizes the validation pipeline. All generated questions were loaded in a lightweight annotation interface and 25.1\% of them (1,092 of \numbergenerated, drawn from every topic and region) were manually validated by two human volunteer annotators, also familiar with the language. For each question the reviewer saw the question, the five options, the correct answer, and the registered explanation. If they could not tell immediately whether the proposed answer was correct, they were free to research the answer with internet resources before giving a verdict: \textit{correct}, \textit{incorrect} or \textit{uncertain}.}

\new{Of the 1,092 reviewed questions, 970 (88.8\%) were judged correct, 79 (7.2\%) incorrect and 43 (3.9\%) uncertain. The most frequent flags were: (i) the question being related to all of France instead of that categorized region; (ii) because of the issue of document-level generation, some of the questions were related to other countries; and (iii) naturalness, meaning that the question was correct but phrased in an unnatural way. Furthermore, the annotators' comments were registered to serve as templates of the intended standard for the automatic filter described in the following section. Annotator instructions are further discussed in Appendix~\ref{ap:Human_ann}.}

\subsection[Automatic Quality Filtering]{\new{Automatic Quality Filtering}}
\label{sec:auto-filter}

\new{The remaining 75\% of questions were screened using an LLM-as-a-judge~\cite{zheng2023judgingneurips}, Claude Opus 5, prompted with the full annotated file, including human verdicts and comments (Appendix~\ref{ap:filter-prompt}). The judge validated questions against human-annotated examples, reformulated those with issues in wording, options, or answers, and assigned a regional specificity score from 1 (unsuitable) to 5 (strongly region-specific). Under a counterfactual criterion, questions whose answers remain unchanged when substituting another French region cannot score 5; well-formed national questions score 4. Human verdicts were preserved verbatim and took precedence over automated judgments. By the end of this step, \numberfiltered\ questions were retained for the difficulty calibration step.}

\subsection[Difficulty Calibration and Second Review]{\new{Difficulty Calibration and Second Validation}}
\label{sec:difficulty}

In \benchmark, question difficulty is defined as the percentage of evaluated LLMs answering correctly at zero-shot (Table~\ref{tab:difficulty_levels}). Higher success rates indicate easier questions, while lower rates indicate greater difficulty (examples from each tier are given in Table~\ref{tab:question_examples_diff}, Appendix~\ref{ap:difficulty}).

\new{All 32 open-weight models in Section~\ref{sec:models} answered the \numberfiltered\ retained questions, classified as \textit{Easy} ($\geq 80\%$ of the models correct), \textit{Medium} (40--80\%), \textit{Hard} (10--40\%) or \textit{suspiciously hard} ($<10\%$, i.e.\ answered incorrectly by over 90\% of the models). The \nSuspicious\ suspiciously hard questions underwent a second volunteer review, including answer research. Reviewers rejected 41 of them, typically due to conflicting statistical sources or multiple defensible answers, and these were removed from the benchmark; the remaining 149 were confirmed and re-labeled \textit{Hard}. After this review, six further questions were re-labeled from \textit{Medium} to \textit{Hard}, giving the final tiers of Table~\ref{tab:difficulty_levels}.}

\begin{table}[t]
\centering
\footnotesize
\setlength{\tabcolsep}{2pt}
\begin{tabular}{@{}llrr@{}}
\toprule
\textbf{Difficulty level} & \textbf{Models correct} & \textbf{Classified} & \textbf{\benchmark} \\
\midrule
Level 1 (Easy)   & $\geq 80\%$ & 808   & 808 \\
Level 2 (Medium) & 40--80\%    & 1,393 & 1,387 \\
Level 3 (Hard)   & 10--40\%    & 782   & 937 \\
Suspicious-hard  & $< 10\%$    & 190   & -- \\
\quad validated $\rightarrow$ Hard & & 149 & \\
\quad rejected $\rightarrow$ removed & & 41 & \\
\midrule
\textbf{Total} & & \textbf{\numberfiltered} & \textbf{\numberquestions} \\
\bottomrule
\end{tabular}
\caption{Difficulty calibration via aggregate 0-shot accuracy of the 32 open-weight models on the \numberfiltered\ questions retained after filtering. Questions that fewer than 10\% of the models answered correctly were flagged as suspicious-hard and sent back to the human reviewers: 149 were confirmed and labeled Hard, 41 were removed. Six further questions were re-labeled from Medium to Hard after the review, giving the final \numberquestions-question benchmark (v1.2).}
\label{tab:difficulty_levels}
\end{table}

Table~\ref{tab:difficulty_levels} shows that the benchmark spans a range of difficulty levels, with a distribution centered on medium difficulty to avoid ceiling and floor effects (see Appendix~\ref{ap:difficulty}).

\section{\Lbench\ Subset}
\label{sec:carte-lv}
We introduce \Lbench, a \new{\nLV-MCQ} subset of \benchmark\ for evaluating regional linguistic variation in France. It probes LLM sensitivity to region-specific vocabulary, expressions, and usage patterns, providing a diagnostic tool for identifying potential biases.

Questions are generated through dedicated prompts covering grammar, discourse markers, pragmatics, register, and regional lexical usage (Appendix~\ref{ap:carte-lv-prompt}). Each includes minimal linguistic context and a regional indication. Since the prompt that generated \Lbench\ is not the same as the other questions also included in the \benchmark\ set, prior to becoming a part of the 4,354 Q\&As, the first question generation of \Lbench\ was pre-filtered with an LLM-as-a-judge considering: (i) grammatical correctness; (ii) linguistic naturalness; (iii) regional appropriateness; and (iv) answer contamination (see more in Appendix~\ref{ap:carte-lv}). \new{The remaining questions are then followed by the benchmark's standard human validation, filtering, and difficulty calibration, reducing the 233 generated items to \nLV\ questions (Table~\ref{tab:questions_per_region_lv}, Appendix~\ref{ap:TopicsDetail}).}

\section{Experimental Setup}

\begin{table*}[t!]
\centering
\footnotesize
\setlength{\tabcolsep}{4pt}
\begin{tabular}{llr rrr rrr rrr r}
\toprule
\textbf{Model} & \textbf{Type} & \textbf{Params} & \multicolumn{3}{c}{\textbf{0-shot}} & \multicolumn{3}{c}{\textbf{1-shot}} & \multicolumn{3}{c}{\textbf{3-shot}} & \textbf{Overall} \\
\cmidrule(lr){4-6}\cmidrule(lr){7-9}\cmidrule(lr){10-12}
 & & & Easy & Med & Hard & Easy & Med & Hard & Easy & Med & Hard & Avg \\
\midrule
\multicolumn{13}{l}{\textit{French-specific LLMs}} \\
Luciole-23B-Instruct & IT & 23.2B & \textbf{99.4} & \textbf{87.7} & \textbf{37.2} & \underline{99.1} & \textbf{86.2} & \textbf{38.7} & \textbf{99.3} & \textbf{86.3} & \textbf{40.3} & \textbf{75.6} \\
Luciole-8B-Instruct & IT & 8.1B & 98.5 & 79.2 & 32.0 & 98.1 & 78.8 & 33.3 & 98.4 & 81.2 & 36.2 & 70.8 \\
Gaperon-1125-8B-SFT & IT & 8.0B & 96.2 & 72.5 & 24.2 & 95.0 & 69.6 & 26.5 & 96.8 & 74.2 & 29.0 & 64.6 \\
Gaperon-1125-8B & Base & 8.0B & 96.2 & 68.0 & 25.5 & 96.4 & 70.7 & 25.3 & 97.9 & 75.9 & 28.3 & 64.5 \\
Claire-Mistral-7B-0.1 & Base & 7.2B & 87.0 & 44.9 & 18.9 & 96.3 & 61.1 & 21.9 & 95.8 & 64.0 & 23.6 & 55.5 \\
Lucie-7B-Instruct-v1.1 & IT & 6.7B & 93.2 & 59.2 & 14.2 & 92.7 & 57.0 & 19.6 & 92.9 & 60.9 & 20.0 & 55.5 \\
Vigogne-2-7B-Instruct & IT & 6.7B & 87.6 & 47.4 & 18.6 & 88.2 & 49.1 & 16.3 & 88.9 & 48.5 & 17.3 & 49.4 \\
Luth-LFM2-1.2B & IT & 1.2B & 85.5 & 46.6 & 20.2 & 82.7 & 47.1 & 22.2 & 82.5 & 47.2 & 23.5 & 48.9 \\
Luciole-1B-Instruct & IT & 1.3B & 76.6 & 38.1 & 19.1 & 80.4 & 44.4 & 18.8 & 81.6 & 45.0 & 17.5 & 44.9 \\
Vigogne-7B-Instruct & IT & 6.7B & 47.6 & 28.6 & 25.1 & 56.3 & 29.6 & 21.8 & 54.5 & 29.3 & 21.9 & 33.4 \\
Claire-7B-0.1 & Base & 6.9B & 27.5 & 20.0 & 20.6 & 23.5 & 20.8 & 20.2 & 22.4 & 19.5 & 21.3 & 21.4 \\
CroissantLLMChat-v0.1 & IT & 1.3B & 25.4 & 20.7 & 17.2 & 19.4 & 21.1 & 20.7 & 20.5 & 21.1 & 19.3 & 20.6 \\
\midrule
\multicolumn{13}{l}{\textit{European multilingual LLMs}} \\
Mistral-Nemo-Instruct-2407 & IT & 12.2B & 98.5 & 79.0 & 27.2 & 98.5 & 82.6 & 31.9 & 98.0 & \underline{84.0} & 33.2 & 70.8 \\
EuroLLM-9B-Instruct-2512 & IT & 9.2B & 99.0 & 78.6 & 20.5 & 98.9 & 81.4 & 25.4 & \underline{99.0} & 80.7 & 25.9 & 68.2 \\
Occiglot-7B-EU5-Instruct & IT & 7.2B & 98.0 & 70.2 & 19.4 & 98.8 & 71.7 & 21.5 & 97.8 & 73.1 & 22.2 & 63.4 \\
Mistral-7B-v0.1 & Base & 7.2B & 98.8 & 65.7 & 17.1 & 98.0 & 68.4 & 19.9 & 96.7 & 68.9 & 19.3 & 60.8 \\
Mistral-7B-Instruct-v0.3 & IT & 7.2B & 98.3 & 66.3 & 16.8 & 96.5 & 65.8 & 17.6 & 95.8 & 65.7 & 14.2 & 59.0 \\
Mistral-7B-Instruct-v0.2 & IT & 7.2B & 94.7 & 61.4 & 14.4 & 93.4 & 57.6 & 14.3 & 92.5 & 59.1 & 14.7 & 54.7 \\
\midrule
\multicolumn{13}{l}{\textit{General multilingual LLMs}} \\
Qwen3.5-9B & IT & 9.0B & \underline{99.3} & \underline{82.2} & 33.0 & \textbf{99.3} & \underline{83.2} & \underline{35.3} & \underline{99.0} & 83.3 & \underline{37.2} & \underline{72.8} \\
Gemma-3-12B-IT & IT & 12.2B & 98.8 & 79.5 & 32.7 & 97.9 & 78.7 & 33.1 & 97.6 & 79.8 & 34.5 & 70.4 \\
GLM-4.7-Flash & IT & 29.9B & 98.3 & 79.7 & \underline{33.1} & 97.2 & 78.2 & 31.2 & 97.6 & 78.1 & 34.0 & 69.8 \\
Llama-3.1-8B-Instruct & IT & 8.0B & \textbf{99.4} & 77.6 & 25.1 & 98.5 & 77.5 & 26.5 & 98.3 & 78.0 & 27.0 & 67.7 \\
Qwen3.5-4B & IT & 4.2B & 97.4 & 76.7 & 27.3 & 96.3 & 75.3 & 26.3 & 96.8 & 73.0 & 27.6 & 66.3 \\
GLM-4-9B-0414 & IT & 9.4B & 98.8 & 72.5 & 25.6 & 98.5 & 72.6 & 29.6 & 97.4 & 75.3 & 28.5 & 66.2 \\
Gemma-4-E4B-IT & IT & 7.9B & 97.3 & 71.2 & 23.6 & 97.8 & 74.2 & 25.2 & 97.4 & 76.1 & 27.1 & 65.4 \\
Aya-Expanse-8B & IT & 8.0B & 99.0 & 70.4 & 24.4 & 96.5 & 67.4 & 26.0 & 97.0 & 69.1 & 26.6 & 63.4 \\
Gemma-4-E2B-IT & IT & 5.1B & 94.7 & 61.2 & 19.2 & 94.3 & 62.5 & 20.8 & 93.1 & 62.3 & 21.3 & 57.8 \\
GLM-4-9B-Chat & IT & 9.4B & 95.9 & 62.4 & 14.7 & 95.7 & 61.8 & 16.2 & 94.6 & 62.3 & 16.5 & 56.9 \\
Llama-3.2-3B-Instruct & IT & 3.2B & 93.2 & 54.5 & 14.9 & 91.6 & 52.8 & 15.6 & 93.3 & 54.3 & 15.4 & 52.3 \\
Gemma-4-E2B & Base & 5.1B & 79.8 & 39.9 & 19.3 & 92.7 & 54.1 & 21.1 & 93.4 & 59.8 & 23.5 & 52.0 \\
Llama-3.2-1B-Instruct & IT & 1.2B & 79.5 & 40.5 & 19.4 & 75.2 & 40.6 & 21.5 & 74.8 & 41.2 & 22.1 & 44.1 \\
BLOOMZ-3B & IT & 3.0B & 81.6 & 38.8 & 20.0 & 46.9 & 22.6 & 14.0 & 54.0 & 28.3 & 20.4 & 34.4 \\
\midrule
\multicolumn{13}{l}{\textit{Frontier API models}} \\
Gemini 3.8 Flash & API & – & 99.3 & 96.6 & 88.8 & 99.9 & 96.5 & 89.0 & 99.8 & 97.3 & 88.8 & 95.2 \\
GLM-5.3-Flash & API & – & 99.4 & 92.1 & 69.8 & 99.3 & 90.9 & 67.6 & 99.0 & 90.1 & 68.6 & 86.5 \\
GLM-5.3 & API & – & 99.5 & 92.2 & 68.5 & 98.5 & 90.1 & 65.0 & 99.3 & 90.8 & 62.3 & 85.4 \\
\bottomrule
\end{tabular}
\caption{Accuracy (\%) on \benchmark\ (\numberquestions\ questions) per difficulty tier and $k$-shot setting ($k \in \{0,1,3\}$) for the 32 open-weight models run locally and for the three frontier API models scored on their generated answer. Type: Base = pretrained checkpoint prompted raw; IT = instruction- or chat-tuned checkpoint prompted through its chat template (BLOOMZ-3B ships no chat template and is prompted raw; the Vigogne models use the template of the \texttt{vigogne} package). The 0-shot columns are computed in float32 (bfloat16 for Luciole-23B-Instruct and GLM-4.7-Flash), the 1- and 3-shot columns in bfloat16. \textbf{Bold} and \underline{underline} denote the highest and second-highest values per column among the open-weight models; \textit{Overall Avg} is the whole-set accuracy averaged over the three shot settings.}
\label{tab:CARTE_results}
\end{table*}

\subsection{Models}
\label{sec:models}

The models were selected to represent both general-purpose and language-specialized models\new{. Thirty-two open-weight models of 1B to 30B parameters were run locally and are analyzed in depth}:

\paragraph{French-specific models.} Models trained primarily on French corpora:
CroissantLLM Chat ($\sim$1.3B)~\cite{faysse2025croissantllm}; Claire-7B and Claire-Mistral-7B~\cite{louradour2024claire}; Lucie-7B-Instruct~\cite{gouvert2025lucie}; Gaperon-1125 (8B, 8B-SFT)~\cite{godey2025gaperon}; Luth-LFM2-1.2B~\cite{lasbordes2026luth}; Vigogne-7B-Instruct / Vigogne-2-7B-Instruct~\cite{vigogne}\new{; and Luciole-1B, -8B and -23B-Instruct~\cite{openllmfrance2026luciole}}.

\paragraph{European multilingual models.} Models with strong French representation in their pretraining corpus: EuroLLM-9B-Instruct (2512 release)~\cite{martins2025eurollm}; Mistral-7B-v0.1, Mistral-7B-Instruct-v0.2, Mistral-7B-Instruct-v0.3, Mistral-Nemo-Instruct-2407~\cite{jiang2023mistral}; and Occiglot 7B \new{EU5} Instruct~\cite{avramidis2024occiglot}.

\paragraph{General multilingual models.} BLOOMZ-3B~\cite{workshop2022bloom}; Gemma-3-12B-IT~\cite{gemmateam2025gemma3}, Gemma-4-E2B\new{, Gemma-4-E2B-IT} and Gemma-4-E4B-IT~\cite{gemmateam2026gemma4}; Llama-3.1-8B-Instruct, Llama-3.2-1B-Instruct, Llama-3.2-3B-Instruct~\cite{grattafiori2024llama3}; Qwen3.5-4B and Qwen3.5-9B~\cite{yang2025qwen3}; Aya Expanse 8B~\cite{dang2024aya}\new{; and GLM-4-9B-0414, GLM-4-9B-Chat and the 30B mixture-of-experts GLM-4.7-Flash~\cite{glm2024chatglm}}.

\paragraph{Frontier API models.} \new{We additionally query three proprietary models through their APIs and score their generated answer: Gemini 3.8 Flash~\cite{team2023gemini}, GLM-5.3 and GLM-5.3-Flash~\cite{glm5team2026glm5}.}

\subsection{Evaluation Protocol}
\label{sec:protocol}

We evaluate each model under 0-shot, 1-shot, and 3-shot settings. \new{Questions follow a fixed French template containing five options and an instruction to select the correct letter. Instruction-tuned models use their chat templates, with the prefix starting the assistant turn, while base models receive plain text. For local models, no text is generated: the predicted answer is the option letter with the highest next-token log-probability; \textit{Je ne sais pas} counts as incorrect. Few-shot demonstrations are randomly sampled (fixed seed) from the benchmark, excluding the target question and its flagged near-duplicates, and formatted as solved user/assistant turns for chat models or concatenated blocks for base models. Accuracy is the primary metric. Zero-shot inference uses float32 weights, except for Luciole-23B-Instruct and GLM-4.7-Flash (bfloat16 due to memory constraints); few-shot inference uses bfloat16. All local experiments run on two 48GB NVIDIA GPUs. BLOOMZ-3B is treated as a base model due to its missing chat template, while Vigogne models use the Alpaca-style template from \texttt{vigogne}. The three API models use the same question template and are evaluated on the letter in their generated responses.}

\section{Results and Discussion}
\label{sec:results}

We summarize our primary evaluation results in Table~\ref{tab:CARTE_results}\new{; per-region results are given in Table~\ref{tab:CartePerRegionAccuracy} (Appendix~\ref{ap:regionresults}) and \Lbench\ results in Table~\ref{tab:LVPerRegionAccuracy} (Appendix~\ref{ap:lvresults})}. Table~\ref{tab:CARTE_results} establishes the baseline capabilities of all models on \benchmark, highlighting the impact of in-context learning and benchmark complexity. \new{Figure~\ref{fig:Best_pre_region} charts the open-weight model that achieves the best accuracy in each region}. A further breakdown of per-topic accuracy is available in Table~\ref{tab:CartePerTopicAccuracy} (Appendix~\ref{ap:regionresults}); abstention rates and option bias per model are given in Table~\ref{tab:behaviour} (Appendix~\ref{ap:behaviour}).

\subsection{Accuracy and Discriminative Capability}

The results show that \benchmark\ effectively distinguishes model pre-training strategies and highlights the competitiveness of French- and culture-specific open-weight models (\textit{Luciole}, \textit{Lucie}, \textit{Vigogne}, and \textit{Gaperon}) relative to their size.

\new{Among the 32 locally evaluated models, \bestModel\ achieves the highest accuracy in every setting (\bestAccZero\% at 0-shot, \bestAccThree\% at 3-shot). Other strong models include Qwen3.5-9B (71.9/73.0/73.6),  Gemma-3-12B-IT (70.5/70.0/70.8), and Luciole-8B-Instruct (70.1/70.2/72.2). French-specific models vary considerably: Claire-7B and CroissantLLM Chat remain near chance, while Vigogne-7B-Instruct barely exceeds it.  Model size correlates with accuracy ($r=0.64$), with an approximately 28-point gain per tenfold increase in parameters (Appendix Figure~\ref{fig:acc_vs_size}). Few-shot prompting generally improves accuracy, but gains vary across models and culturally specific subsets.}


\new{Frontier API models perform better still: Gemini 3.8 Flash reaches about 95\% and the two GLM-5.3 models 85--87\% (Table~\ref{tab:CARTE_results}). However, the questions were generated by an earlier model of the same family (Gemini 3 Flash), so familiarity with its own formulations may partly explain this advantage.} \rev{The benchmark is not saturated for its target models: the best open-weight model below 30B answers roughly three quarters of questions, French-specific models answer about half, and no local model exceeds 41\% on the human-validated Hard tier (Section~\ref{sec:complexity}).}


\subsection{Benchmark Complexity and Difficulty}
\label{sec:complexity}



While most models perform well on the Easy tier, performance declines on Hard questions, highlighting the challenge of region-specific nuances even for large models. Overall, locally run models peak at approximately \new{76}\% average accuracy. Easy-tier accuracy exceeds 80\% for most models, except the two chance-level models, Vigogne-7B-Instruct, and three slightly below the threshold (Luciole-1B-Instruct, Llama-3.2-1B-Instruct, and base Gemma-4-E2B). Medium questions generally benefit from few-shot prompting, whereas Hard questions show greater variability, sometimes degrading with additional shots, likely due to irrelevant demonstrations introducing noise.

\new{On the Hard tier, no local model exceeds 41\% accuracy: Luciole-23B-Instruct leads with 40.3\% (3-shot) and 37.2\% (0-shot), followed by Qwen3.5-9B, Luciole-8B-Instruct, Gemma-3-12B-IT, and GLM-4.7-Flash (31–37\%). Although low accuracy is expected because this tier is defined by model failures, a second human verification of all 149 suspiciously hard questions supports their validity and answerability rather than annotation noise.}

\section{Conclusion}

In this paper, we introduced \benchmark\ (\textbf{C}ulturally \textbf{A}nchored \textbf{R}egional-\textbf{T}erritorial \textbf{E}valuation) and its linguistic subset \Lbench, addressing the lack of regionally grounded French evaluation benchmarks. Our validation pipeline combines human review, an LLM-as-a-judge filter, model-based difficulty calibration, and a second human review.

Evaluation of \new{\numbermodels} open-weight and three API models reveals substantial gaps in regional knowledge and linguistic variation. Although overall accuracy may appear high, performance varies across regions and topics: the best open-weight model below 30B parameters answers 75\% of the questions, accuracy differs by twelve points between the best- and least-known regions, and several models show marked answer-position biases.

\new{Future work includes extending human validation to the full benchmark, covering overseas territories, and investigating whether regional pretraining data or retrieval can reduce the identified gaps.}

\label{sec:conclusion}

\section*{Limitations}
\label{sec:limitations}

Questions for \benchmark\ and \Lbench\ were generated automatically from curated sources. Although 25\% were human-annotated and the remainder screened by an LLM judge calibrated on these annotations, undetected errors may persist.

Regional coverage is uneven due to source availability, while model selection is constrained by computational resources and open-weight model availability. Source materials and automated generation may also introduce biases, reinforcing stereotypes or privileging dominant narratives over minority, local, or evolving perspectives.

\rev{The benchmark covers only metropolitan France's 13 regions, excluding the distinct linguistic and cultural contexts of the overseas regions (Guadeloupe, Martinique, Guyane, La Réunion, Mayotte) and cross-border Francophone territories (Wallonia, Romandy), which are priorities for future versions.}

\section*{Ethical Considerations}
\benchmark\ and \Lbench\ are intended for non-commercial research on regional reasoning in LLMs, not for profiling populations or supporting political, normative, or sociocultural judgments about French territories. All sources are publicly accessible and used for factual information without intentional text reproduction. Volunteers participated knowingly and without payment, using an annotation interface that stored no personal data.

We acknowledge that geographically grounded datasets may reproduce regional stereotypes. To mitigate this risk, questions rely exclusively on public statistical, geographical, and historical sources and emphasize verifiable facts over subjective cultural judgments.

The benchmarks are not intended for high-stakes use; scores should not be interpreted as measures of real-world regional competence or cultural sensitivity. A ``Je ne sais pas'' option also allows uncertainty rather than forced guessing.

\section*{Acknowledgements}

We thank Inès Benito, Enzo Pinchon, Geoffrey Deperle, and Dr. Johannes Lutzeyer for their support during early benchmark verification and for insightful discussions on unnatural constructions in region-specific sentence generation and future benchmark improvements.  This work has benefited from the support of the OpenLLM France project, funded by Bpifrance as part of the France 2030 program “Communs numériques pour l’intelligence artificielle générative”.

\bibliography{carte}
\bibliographystyle{acl_natbib}

\newpage
\appendix

\section{\benchmark\ Question Generation Prompt}
\label{ap:MCQprompt}

The following text is the prompt used for the generation of the questions used in \benchmark\ (question examples from each difficulty tier are given in Table~\ref{tab:question_examples_diff}):

\begin{quote}
\small

\texttt{\textbf{RÔLE:} Assume le rôle d’un spécialiste de la France : économie, traditions, langue, transports, biodiversité, démographie, et tout ce qui concerne la France et la langue française.}\\[2pt]

\texttt{\textbf{ENTRÉE:} À partir des informations contenues dans le document fourni:}\\[2pt]

\texttt{<DOCUMENT>}\\[2pt]

\texttt{\textbf{OBJECTIF:} Identifier le principal objectif du document en extrayant uniquement ses thèmes centraux. À partir de ces éléments, générer 30 questions à choix multiples (QCM) en lien avec la France ou la langue française, en couvrant, lorsque pertinent, les thématiques suivantes :}\\[2pt]

      \texttt{*Les dynamiques économiques régionales;}\\
      \texttt{*Les infrastructures et réseaux de transport;}\\
      \texttt{*Les caractéristiques démographiques;}\\
      \texttt{*Les spécificités culturelles et linguistiques}\\
      \texttt{*La biodiversité et les territoires;}\\
      \texttt{*Les institutions, traditions et réalités sociales françaises;}\\[2pt]

\texttt{\textbf{CONTRAINTE FONDAMENTALE:} Le document est uniquement une source de contenu. Il est strictement interdit de :}\\[2pt]

      \texttt{*Mentionner le document ou le lien}\\
      \texttt{*Faire référence à sa structure, ses sections ou sa formulation}\\
      \texttt{*Utiliser des expressions telles que « selon le document », « dans le texte », « dans le lien »}\\
      \texttt{*Construire des questions dépendant de la formulation ou de l’organisation du document}\\
    \texttt{*Toutes les questions doivent être reformulées comme des connaissances générales indépendantes.}\\
    \texttt{*Le document est la pour le contenu, mais ne doit jamais être cité ni requis pour répondre.}\\[2pt]

\texttt{\textbf{EXIGENCES DES QUESTIONS:} Chaque question doit :}\\
      \texttt{*être totalement autonome et compréhensible sans aucune source}\\
      \texttt{*être fondée sur des connaissances réelles en économie, géographie ou démographie}\\
      \texttt{*inclure obligatoirement un contexte territorial explicite}\\
      \texttt{*éviter toute ambiguïté et reposer sur un raisonnement clair}\\
\texttt{Chaque question doit inclure un ancrage territorial explicite, par exemple :}\\
      \texttt{*une région française}\\
      \texttt{* un espace géographique clairement défini}\\
\texttt{Les questions sans contexte  régional sont interdites.}\\[2pt]

\texttt{\textbf{CONTRAINTE TEMPORELLE (OBLIGATOIRE):} Si une question porte sur :}\\[2pt]
      \texttt{*Une tendance économique}\\
      \texttt{*Une politique publique}\\
      \texttt{*Une infrastructure}\\
      \texttt{*Une évolution démographique}\\
      \texttt{*Un changement historique}\\
      \texttt{*Une réforme ou transformation structurelle}\\
\texttt{Alors elle doit inclure une référence temporelle explicite.}\\[2pt]

\texttt{Formats autorisés :}\\[2pt]
      \texttt{« En 2020 »}\\
       \texttt{« Depuis 2010 »}\\
      \texttt{« Dans les années 1990 »}\\
      \texttt{« À partir de 2015 »}\\
      \texttt{« Au début des années 2000 »}\\[2pt]

\texttt{Formats interdits :}\\[2pt]
      \texttt{« Récemment »}\\
      \texttt{« Aujourd’hui »}\\
      \texttt{« Actuellement »}\\
      \texttt{« Historiquement » sans précision de date}\\[2pt]

\texttt{\textbf{QUALITÉ DES QUESTIONS:} Les questions doivent être :}\\[2pt]
      \texttt{*Non triviales}\\
      \texttt{*Analytiques}\\
      \texttt{*Fondées sur des dynamiques territoriales réelles}\\
      \texttt{*Variées dans leur structure et leur raisonnement (causalité, comparaison, analyse spatiale, interprétation)}\\
      \texttt{*Indépendantes de toute référence à un document}\\[2pt]

\texttt{\textbf{RÈGLES DES DISTRACTEURS:} Chaque question comporte 5 choix (A à E). Règles :}\\
      \texttt{*A à D sont plausibles mais une seule réponse est correcte;}\\
      \texttt{*E est toujours « Je ne sais pas »;}\\
      \texttt{*Les distracteurs doivent être réalistes et géographiquement crédibles;}\\
      \texttt{*Ils doivent appartenir à d’autres régions françaises ou européennes;}\\
      \texttt{*Il ne doit exister qu’une seule réponse correcte, mais les distracteurs doivent être non triviaux et difficiles;}\\
\texttt{Interdits :}\\
      \texttt{*Réponses vagues;}\\
      \texttt{*Ambiguïté entre plusieurs bonnes réponses;}\\
      \texttt{*Distracteurs non réalistes ou imprécis;}\\
      \texttt{*Chevauchement territorial confus;}\\

\texttt{\textbf{VARIATION OBLIGATOIRE:} Garantir une forte diversité dans les questions afin d’éviter tout effet de répétition ou de monotonie. Concrètement :}\\

      \texttt{*éviter d’enchaîner plusieurs questions de même nature (par exemple uniquement des questions chiffrées ou statistiques);}\\
      \texttt{*alterner systématiquement les formats : localisation (“où ?”), explication (“pourquoi ?”), comparaison (“quel est le plus… ?”), identification (“quel élément… ?”), analyse de situation;}\\
      \texttt{*varier les échelles géographiques (commune, région, espace rural/urbain, national);}\\
      \texttt{*ne pas concentrer les questions sur un seul type d’infrastructure (ex : uniquement routes ou uniquement trains);}\\
      \texttt{*diversifier les angles d’approche pour un même thème (ex : un port peut être abordé par son rôle économique, son trafic, ou sa localisation stratégique) ;}\\
      \texttt{*éviter les suites de questions portant sur le même secteur économique (ne pas faire plusieurs questions consécutives sur l’agriculture, par exemple);}\\
      \texttt{*varier les niveaux cognitifs : connaissances simples, compréhension, mise en relation, interprétation de situations;}\\
      \texttt{*Éviter toute répétition excessive de structure ou de région;}\\
\end{quote}

\section{\Lbench\ Question Generation Prompt}
\label{ap:carte-lv-prompt}

The following text is the prompt used for the generation of the questions used in \Lbench:

\begin{quote}

\small
\texttt{\textbf{RÔLE:}Vous êtes un expert des variations linguistiques à travers les régions françaises.}\\[2pt]

\texttt{\textbf{ENTRÉE:} À partir des informations
contenues dans le document fourni:}\\[2pt]

\texttt{<DOCUMENT>}\\[2pt]

\texttt{\textbf{OBJECTIF:} Générer 30 questions à choix multiples (QCM) en français portant sur les variations linguistiques régionales en France. Les questions doivent couvrir l’ensemble des 13 régions françaises et évaluer la capacité à reconnaître, interpréter ou comparer des usages linguistiques régionaux authentiques.}\\[2pt]

\texttt{Les questions doivent être fondées sur des phénomènes linguistiques observables dans l’usage réel :}\\
\texttt{- lexique régional;}\\
\texttt{- variations grammaticales;}\\
\texttt{- marqueurs discursifs;}\\
\texttt{- différences pragmatiques;}\\
\texttt{- variations de registre (oral vs écrit);}\\
\texttt{- prononciations représentées à l’écrit;}\\
\texttt{- expressions idiomatiques régionales;}\\
\texttt{- usages conversationnels contextualisés;}\\

\texttt{L’objectif est d’évaluer la sensibilité aux variations régionales du français à travers des situations naturelles et linguistiquement plausibles.}\\[2pt]

\texttt{\textbf{CONTRAINTE FONDAMENTALE:} Une question valide doit obligatoirement contenir:}\\

\texttt{- un contexte d’usage minimal réaliste;}\\
\texttt{- un signal linguistique observable;}\\
\texttt{- un contraste explicite ou implicite entre plusieurs formes possibles;}\\

\texttt{Le contraste peut porter sur :}\\
\texttt{- deux structures grammaticales;}\\
\texttt{- deux choix lexicaux concurrents;}\\
\texttt{- deux marqueurs discursifs;}\\
\texttt{- deux registres (oral vs écrit);}\\
\texttt{- deux formulations pragmatiques;}\\
\texttt{- une variante régionale vs une forme standard;}\\[2pt]

\texttt{\textbf{INTERDICTIONS:}NE JAMAIS générer:}\\

\texttt{- des questions sans situation d’usage;}\\
\texttt{- des formulations abstraites sans ancrage conversationnel;}\\
\texttt{- des régionalismes inventés;}\\
\texttt{- des dialectes artificiels;}\\
\texttt{- plusieurs réponses peuvent être interprétées comme correctes;}\\[2pt]

\texttt{\textbf{EXIGENCES DES QUESTIONS:} Chaque question doit:}\\

\texttt{- être rédigée entièrement en français;}\\
\texttt{- inclure une région française explicite;}\\
\texttt{- contenir un micro-contexte réaliste :
  dialogue, interaction sociale, école, famille, marché, sport, café, administration, etc;}\\
\texttt{- tester un phénomène linguistique authentique;}\\
\texttt{- inclure 5 options :
  A, B, C, D, E = « Je ne sais pas »;}\\
\texttt{- avoir une seule bonne réponse;}\\
\texttt{- proposer des distracteurs plausibles mais incorrects;}\\
\texttt{- rester naturelle et crédible dans l’usage oral ou écrit;}\\
\texttt{- refléter des usages attestés régionalement;}\\[2pt]

\texttt{\textbf{VARIATION OBLIGATOIRE:} Sur l’ensemble des questions générées :}\\

\texttt{- varier les structures syntaxiques;}\\
\texttt{- alterner : lexique régional, pragmatique, grammaire, discours, registres, expressions idiomatiques, prononciation représentée à l’écrit;}\\
\texttt{- éviter les répétitions de patrons de questions}\\[2pt]
\end{quote}

\section[Validation and Filtering Prompt]{\new{Validation and Filtering Prompt}}
\label{ap:filter-prompt}

\new{The LLM judge of Section~\ref{sec:auto-filter} receives the annotated question file (question, options, correct answer, explanation, region, topic) and the human annotators' \texttt{verdicts}, \texttt{flags}, \texttt{issues} and \texttt{notes} for 25.1\% of the questions. The following instructions were used for filtering:}

\begin{quote}
\small
\texttt{\textbf{Objective.} You are filtering and improving a benchmark of questions about French regional culture. For each question, determine how well it measures regional cultural knowledge for the region specified in its annotation. The key criterion is regional specificity: a question receives a high \texttt{local\_score} only when answering it requires knowledge genuinely specific to the annotated region, rather than knowledge that applies equally to France as a whole. Merely mentioning the name of a region is not sufficient. Perform three tasks: (1) validate the question; (2) improve or reformulate it when necessary; (3) assign a \texttt{local\_score} from 1 to 5 and filter the final dataset.}\\[3pt]

\texttt{\textbf{Human-validated questions.} A human verdict may be \texttt{correct}, \texttt{incorrect} or \texttt{uncertain}. Human validation takes precedence over automatic validation. \texttt{uncertain} means the annotator identified something potentially problematic; inspect the wording, factual basis, answer choices, correct answer and regional relevance, and determine whether the question can be corrected based on the annotations or on similar examples. Do not assume that an uncertain question is incorrect.}\\[3pt]

\texttt{\textbf{Questions without a human verdict.} Evaluate the question yourself; determine whether it is factually and linguistically appropriate; infer the intended standard from similar questions that already have human annotations; reformulate if necessary; assign the \texttt{local\_score} and an automatic verdict (\texttt{automatic-correct} only if the question is explicitly determined to be correct, otherwise \texttt{automatic-incorrect}). The absence of a human verdict does not mean the question is correct.}\\[3pt]

\texttt{\textbf{Scores.} 5 -- strongly region-specific: the knowledge tested is genuinely characteristic of the target region; a person who only knows general French culture should not be able to answer reliably. 4 -- valid and culturally relevant to France but not specific to the sub-national region (national institution, France-wide practice, or \texttt{region = France}). 3 -- weak, indirect or incidental regional connection. 2 -- mostly non-regional: the answer depends on general or international knowledge. 1 -- unsuitable: unrelated, malformed, unanswerable, wrong regional association, or too specific to a local business or person with no regional relevance.}\\[3pt]

\texttt{\textbf{Regionality rule.} If the target region were replaced by another French region, would the question remain equally valid with the same answer? If yes, it is probably not a 5. Could someone answer correctly using only general knowledge about France? If yes, assign 4 or below.}\\[3pt]

\end{quote}

\section{\Lbench\ Filtering}
\label{ap:carte-lv}

Each question is independently evaluated using two models, Gemini and ChatGPT, along four criteria:

\begin{itemize}
    \item \textbf{Grammatical Correctness (GC):} structural correctness according to standard French grammar.
    \item \textbf{Linguistic Naturalness (LN):} degree of fluency and idiomaticity as perceived by a native speaker, beyond strict grammatical rules.
    \item \textbf{Regional Appropriateness (RA):} suitability of the expression for the intended Francophone region and context; when no region is specified, standard metropolitan French is assumed.
    \item \textbf{Answer Contamination (AC):} presence of cues within the question that explicitly or implicitly reveal the correct answer.
\end{itemize}

Each question is assigned a score on a 1–5 scale for each criterion. We retain only questions scoring 4 or higher, and discard all remaining items to ensure high-quality and consistent annotations across the dataset. Scoring Scale (1–5) follows:

Score, 1, when:
\begin{itemize}
    \item GC: Major grammatical errors; difficult to understand
    \item LN: Unnatural or non-native-like phrasing
    \item RA: Not used or inappropriate in context
    \item AC: Strongly contains or reveals the answer
\end{itemize}

Score, 2, when:
\begin{itemize}
    \item GC: Frequent grammar issues
    \item LN: Understandable but awkward
    \item RA: Rare or odd in context
    \item AC: Some hint of answer included
\end{itemize}

Score, 3, when:
\begin{itemize}
    \item GC: Mostly correct with minor issues
    \item LN: Neutral but not idiomatic
    \item RA: Plausible but not typical
    \item AC: Slight leakage or ambiguity
\end{itemize}

Score, 4, when:
\begin{itemize}
    \item GC: Correct with minor or negligible issues
    \item LN: Natural and fluent
    \item RA: Appropriate for context/region
    \item AC: No meaningful answer leakage
\end{itemize}

Score, 5, when:
\begin{itemize}
    \item GC: Fully correct grammar
    \item LN: Fully native-like and idiomatic
    \item RA: Fully natural for the specified region/context
    \item AC: Completely neutral, no answer implied
\end{itemize}

After balancing the dataset we see in Table~\ref{tab:questions_per_region_lv} the distributions of questions per regions for \Lbench.

\section{Topic Detail}
\label{ap:TopicsDetail}
\label{ap:qpc}

Here, we provide a more detailed breakdown of the subtopics grouped under each \new{of the 12} defined topic\new{s}.

\begin{itemize}
    \item Institutions \& gouvernance: This category covers the structure, functioning, and evolution of public and administrative institutions across multiple scales (local, national, and European). It includes themes related to political and administrative organization, public policy, governance systems, institutional history, legal frameworks, and state reform. It also encompasses the interaction between institutions and broader societal dimensions such as territory, economy, culture, education, technology, and social realities, as well as cross-cutting topics including laïcité, European integration, public services, and institutional values.

    \item Économie \& industrie: This category covers economic systems, labor markets, and industrial structures, including employment dynamics, workforce training, and evolving forms of work. It encompasses both macroeconomic and sectoral perspectives such as industrial production, strategic industries, innovation ecosystems, energy and digital economies, and financial mechanisms. The category further includes themes related to economic policy, investment, trade, and territorial economic development, as well as competitiveness, attractiveness, and international influence. It also addresses industry-specific domains (e.g., aerospace, pharmaceutical, and manufacturing sectors), industrial transformation processes such as reindustrialization and deindustrialization, and the interplay between economic activity and broader societal, environmental, technological, and infrastructural factors.

    \item Agriculture \& terroirs: This category encompasses agricultural systems, practices, and rural dynamics, including production methods, agricultural development, and the organization of farming activities. It also covers the interaction between agriculture and environmental factors such as landscapes, biodiversity, and land use. A key focus is placed on the notion of \textit{terroirs}, including local traditions, culinary heritage, gastronomy, and the cultural identity of rural regions. The category further includes economic dimensions of agriculture, such as agri-food value chains, protection of regional products, and the role of agriculture in local and national economies, as well as its connections to language, heritage, and traditional knowledge systems.

    \item Infrastructure \& réseaux: This category covers physical, digital, and organizational infrastructures that support territorial cohesion and economic and social activity. It includes transport infrastructure (road, rail, maritime, and air networks), communication systems, and digital infrastructures such as broadband and information networks. It also encompasses energy infrastructure and utility systems, as well as public service facilities including educational, sporting, and administrative infrastructures. A further focus is placed on the role of infrastructure in territorial planning and development, including its interactions with industry, environment, and service provision, as well as its historical evolution and contribution to accessibility and regional connectivity.

    \item Transport \& mobilité: This category focuses on the systems, practices, and policies governing the movement of people, goods, and services across territories. It includes daily mobility patterns, transportation networks, and multimodal transport systems such as road, rail, air, and maritime infrastructure. It also covers logistics and supply chain organization, transport safety (including road safety), and the integration of transport systems with territorial planning and economic activity. In addition, the category addresses the interactions between transport and broader societal and environmental dimensions, including sustainability, demographic dynamics, technological developments in mobility, and the historical evolution of transport systems.

    \item Environnement \& biodiversité: This category encompasses natural systems, ecological processes, and environmental dynamics, including biodiversity, ecosystems, and species conservation. It covers environmental science topics such as climate systems, natural hazards, hydrology, geology, soil and land use, and the management of natural resources. The category also includes environmental governance and policy, such as environmental law, climate action, and sustainability transitions, as well as interactions between ecosystems and human activities, including agriculture, infrastructure, transport, and economic development. Particular attention is given to territorial and coastal environments, environmental heritage, and the historical evolution of human–environment relationships.

    \item Territoire \& aménagement: This category covers spatial organization, territorial identity, and the planning and development of geographic areas at multiple scales. It includes themes related to territorial governance, administrative geography, spatial analysis, and regional dynamics, as well as urban and rural planning, housing, and infrastructure distribution. The category also encompasses urban development processes, architecture, and land-use planning, including the evolution of cities and metropolitan areas. In addition, it addresses the relationships between territories and broader socio-economic, environmental, cultural, and linguistic factors, as well as issues related to regional comparison, territorial inequality, risk management, and heritage within spatial contexts.

    \item Société \& réalités sociales: This category encompasses the study of social structures, practices, and dynamics that shape everyday life in French society. It includes themes related to social organization, cultural norms, values, and traditions, as well as key societal domains such as health, housing, family life, and living conditions. The category also covers social change and inequality, including migration, demographic evolution, labor conditions, and reform processes. In addition, it addresses the interactions between society and other dimensions such as politics, territory, economy, religion, education, and culture, with a focus on sociological perspectives on urban life, social practices, and collective behavior.

    \item Démographie: This category covers the statistical and structural analysis of population dynamics, including population size, distribution, composition, and evolution over time. It encompasses demographic characteristics at national, regional, and local levels, as well as their interactions with social, economic, administrative, and territorial factors. The category also includes population-related phenomena such as migration, education, employment, housing, and mobility, along with sector-specific dimensions such as school and labor force demographics. In addition, it addresses the relationships between demographic trends and broader societal systems, including public services, policy planning, spatial organization, and historical or linguistic variations.

    \item Langue \& sciences du langage: This category encompasses the study of language structure, use, variation, and evolution, with a particular focus on the French language and its regional, social, and historical dimensions. It includes core linguistic subfields such as phonetics, sociolinguistics, dialectology, semantics, and etymology, as well as applied linguistics and language perception. The category also addresses language change over time, linguistic diversity, and regional or social variation, including issues of inclusion, identity, and territorial specificity. In addition, it explores the interactions between language and other domains such as education, culture, history, agriculture, and demographic or institutional contexts.

    \item Histoire \& patrimoine: This category covers historical developments, cultural heritage, and the processes through which collective memory and identity are constructed and preserved. It includes the study of French, regional, and European history, as well as historical geography and long-term territorial transformations. The category also encompasses tangible and intangible heritage, including architectural, cultural, musical, and artisanal traditions, as well as UNESCO-recognized knowledge and savoir-faire. In addition, it addresses the relationship between history and other domains such as institutions, environment, economy, tourism, and spatial planning, highlighting how historical processes shape contemporary territorial and cultural landscapes.

    \item Culture, traditions \& société: This category encompasses cultural practices, symbolic systems, and collective representations that shape social identity and cohesion across French territories. It includes traditions, regional cultures, gastronomy, and culinary heritage, as well as artistic, literary, musical, and architectural expressions. The category also covers cultural dynamics such as cultural diffusion, influence, and transmission, including both historical and contemporary perspectives. In addition, it addresses the relationship between culture and other societal dimensions, including language, religion, institutions, territory, and technology, with particular attention to regional and urban cultural specificities. It further includes intangible cultural heritage such as folklore, myths, artisanal practices, and ways of life, as well as contemporary cultural phenomena such as digital culture and cultural globalization. Finally, it integrates perspectives on identity, cultural diversity, and the role of traditions in shaping social, economic, and environmental interactions.
\end{itemize}

\new{Table~\ref{tab:questions_region_category} gives the number of questions of the \benchmark\ per category and region. Table \ref{tab:questions_per_region_lv} the number of questions of \Lbench\ per region.}

\begin{table*}[htb!]
\centering
\resizebox{\textwidth}{!}{%
\begin{tabular}{lrrrrrrrrrrrrrrr}
\toprule
\textbf{Category} & \textbf{ARA} & \textbf{BFC} & \textbf{BRE} & \textbf{CVL} & \textbf{COR} & \textbf{FR} & \textbf{GE} & \textbf{HdF} & \textbf{NOR} & \textbf{NA} & \textbf{OCC} & \textbf{PdL} & \textbf{PACA} & \textbf{IdF} & \textbf{Total} \\
\midrule
Agriculture \& terroirs & 17 & 34 & 14 & 1 & 4 & 1 & 13 & 9 & 10 & 16 & 12 & 5 & 36 & 1 & 173 \\
Culture, traditions \& société & 40 & 44 & 35 & 17 & 23 & 37 & 46 & 44 & 38 & 45 & 26 & 15 & 59 & 75 & 544 \\
Démographie & 34 & 29 & 9 & 21 & 1 & 14 & -- & 8 & 11 & 17 & 14 & 14 & 17 & 15 & 204 \\
Environnement \& biodiversité & 41 & 27 & 13 & 4 & 17 & 5 & 32 & 20 & 19 & 26 & 18 & 6 & 41 & 2 & 271 \\
Histoire \& patrimoine & 76 & 67 & 38 & 16 & 13 & 11 & 50 & 41 & 35 & 38 & 43 & 20 & 74 & 82 & 604 \\
Infrastructure \& réseaux & 16 & 26 & 7 & 5 & 1 & 1 & 19 & 5 & 12 & 18 & 14 & 16 & 11 & 16 & 167 \\
Institutions \& gouvernance & 12 & 16 & 11 & 5 & 11 & 26 & 16 & 15 & 18 & 13 & 9 & 9 & 12 & 41 & 214 \\
Langue \& sciences du langage & 22 & 21 & 15 & 21 & 19 & 24 & 23 & 21 & 11 & 21 & 29 & 15 & 14 & 21 & 277 \\
Société \& réalités sociales & 9 & 12 & 4 & 9 & 3 & 11 & 10 & 8 & 6 & 9 & 9 & 13 & 18 & 34 & 155 \\
Territoire \& aménagement & 17 & 8 & 10 & 7 & 11 & 1 & 4 & 6 & 7 & 7 & 7 & 3 & 12 & 7 & 107 \\
Transport \& mobilité & 1 & 1 & 20 & 13 & -- & 1 & 1 & 2 & 1 & 3 & 1 & 3 & 7 & 4 & 58 \\
Économie \& industrie & 42 & 44 & 24 & 18 & 4 & 9 & 32 & 11 & 36 & 27 & 30 & 25 & 30 & 26 & 358 \\
\midrule
\textbf{Total} & \textbf{327} & \textbf{329} & \textbf{200} & \textbf{137} & \textbf{107} & \textbf{141} & \textbf{246} & \textbf{190} & \textbf{204} & \textbf{240} & \textbf{212} & \textbf{144} & \textbf{331} & \textbf{324} & \textbf{3,132} \\
\bottomrule
\end{tabular}%
}
\caption{Number of questions per category and region in \benchmark. ARA: Auvergne-Rhône-Alpes, BFC: Bourgogne-Franche-Comté, BRE: Bretagne, CVL: Centre-Val de Loire, COR: Corse, FR: France (national scope), GE: Grand Est, HdF: Hauts-de-France, NOR: Normandie, NA: Nouvelle-Aquitaine, OCC: Occitanie, PdL: Pays de la Loire, PACA: Provence-Alpes-Côte d'Azur, IdF: Île-de-France.}
\label{tab:questions_region_category}
\end{table*}

\begin{table}[htb!]
\centering
\begin{tabular}{lr}
\toprule
\textbf{Region} & \textbf{\# Questions} \\
\midrule
Grand Est & 21 \\
Occitanie & 21 \\
Auvergne-Rhône-Alpes & 20 \\
Bourgogne-Franche-Comté & 20 \\
Hauts-de-France & 20 \\
Corse & 18 \\
Nouvelle-Aquitaine & 18 \\
Centre-Val de Loire & 17 \\
Île-de-France & 17 \\
Pays de la Loire & 15 \\
Bretagne & 11 \\
Normandie & 11 \\
Provence-Alpes-Côte d'Azur & 11 \\
\midrule
\textbf{Total} & \textbf{\nLV} \\
\bottomrule
\end{tabular}
\caption{Number of questions per region (\Lbench, after validation and filtering).}
\label{tab:questions_per_region_lv}
\end{table}

\section[Human Annotation Instructions]{\new{Human Annotation Instructions}}
\label{ap:Human_ann}

\new{The following instructions describe the voluntary and unpaid participation process for annotators responsible for evaluating multiple-choice questions designed to assess large language models’ (LLMs’) knowledge of regional, cultural, and linguistic variations in France. Annotators are provided with an HTML interface containing questions, five answer options (A–E), a proposed correct answer, and an explanation supporting that answer. They are expected to carefully review each question, assess whether it is clearly formulated and relevant to the benchmark, and verify the factual accuracy of the proposed answer and its explanation. When the provided explanation is unclear or unconvincing, annotators are encouraged to consult reliable external sources to establish the correct answer. They should use the comments field to document any issues that could affect a question’s validity, including ambiguity, factual inaccuracies, multiple plausible answers, inadequate explanations, or questionable cultural and linguistic assumptions. Finally, annotators must select “Validate” for questions that are sufficiently clear, relevant, and supported by a defensible answer, or “Reject” when significant issues compromise their validity, providing a concise justification for each rejection. Throughout the process, decisions should be based on objective evidence and sound reasoning, while accounting for the diversity of regional practices, cultural traditions, and linguistic expressions across France.

\begin{quote}
\small

\texttt{\textbf{INSTRUCTIONS:} La participation à ce processus d’annotation est entièrement volontaire. En y participant, vous acceptez de contribuer à l’évaluation de questions visant à mesurer les connaissances des grands modèles de langue (LLM) concernant les variations régionales, culturelles et linguistiques en France.\\[4pt] \textbf{Veuillez noter que cette activité est bénévole et qu’aucune rémunération ne sera versée en contrepartie de votre participation.}}\\[4pt]

\texttt{\textbf{ENTRÉE :} Vous aurez accès à une interface HTML contenant des questions à choix multiples (QCM), chacune comprenant cinq options (A à E), une réponse indiquée comme correcte et une explication justifiant cette réponse.}\\[4pt]

\texttt{\textbf{OBJECTIF :} Évaluer la validité de chaque question en vérifiant sa clarté, sa pertinence et l’exactitude de la réponse indiquée comme correcte.}\\[4pt]

\texttt{\textbf{PROCÉDURE D’ÉVALUATION :}}\\[2pt]

\texttt{*Lire attentivement la question et les cinq options de réponse ;}\\
\texttt{*Vérifier que la question est claire, compréhensible et pertinente au regard des variations régionales, culturelles ou linguistiques en France ;}\\
\texttt{*Vérifier l’exactitude de la réponse indiquée comme correcte et évaluer si l’explication fournie est suffisamment convaincante ;}\\
\texttt{*En cas de doute, consulter des sources externes fiables, notamment des ressources en ligne, afin de vérifier les informations ;}\\
\texttt{*Signaler dans le champ de commentaires tout problème susceptible de compromettre la validité de la question ;}\\
\texttt{*Sélectionner « Validate » ou « Reject » pour enregistrer la décision finale.}\\[4pt]

\texttt{\textbf{CRITÈRES DE VALIDATION :} Une question doit être validée lorsque sa formulation est suffisamment claire, qu’elle est pertinente pour le benchmark et que la réponse indiquée comme correcte est factuellement défendable.}\\[4pt]

\texttt{\textbf{CRITÈRES DE REJET :} Une question doit être rejetée lorsque des problèmes importants compromettent sa validité, notamment :}\\[2pt]

\texttt{*Une réponse indiquée comme correcte qui est factuellement erronée ou insuffisamment justifiée ;}\\
\texttt{*Une formulation ambiguë, confuse ou trompeuse ;}\\
\texttt{*Plusieurs réponses plausibles sans possibilité de distinguer clairement la réponse attendue ;}\\
\texttt{*L’absence de réponse clairement correcte ;}\\
\texttt{*Une explication incorrecte, incomplète ou trompeuse ;}\\
\texttt{*Un contenu non pertinent pour les objectifs du benchmark ou reposant sur des généralisations culturelles ou linguistiques discutables.}\\[4pt]

\texttt{\textbf{COMMENTAIRES :} Utilisez le champ de commentaires pour décrire précisément les problèmes identifiés et, lorsque cela est possible, fournir des éléments factuels ou des références permettant de justifier votre décision. Tout rejet doit être accompagné d’une justification concise.}\\[4pt]

\texttt{\textbf{PRINCIPES FONDAMENTAUX :} Fondez vos décisions sur des faits vérifiables et un raisonnement objectif, plutôt que sur vos préférences personnelles. Tenez compte de la diversité des pratiques, des expressions et des réalités culturelles selon les régions et les communautés. Ne rejetez pas une question uniquement parce que le fait présenté vous est inconnu.}\\[4pt]

\texttt{\textbf{DÉCISION FINALE :} Sélectionnez « Valider » si la question est suffisamment claire, pertinente et possède une réponse défendable. Sélectionnez « Rejeter » si des problèmes importants compromettent sa validité.}

\end{quote}}

\section[Results per Region]{\new{Results per Region}}
\label{ap:regionresults}

\begin{figure*}[t!]
  \centering
  \includegraphics[width=1\textwidth]{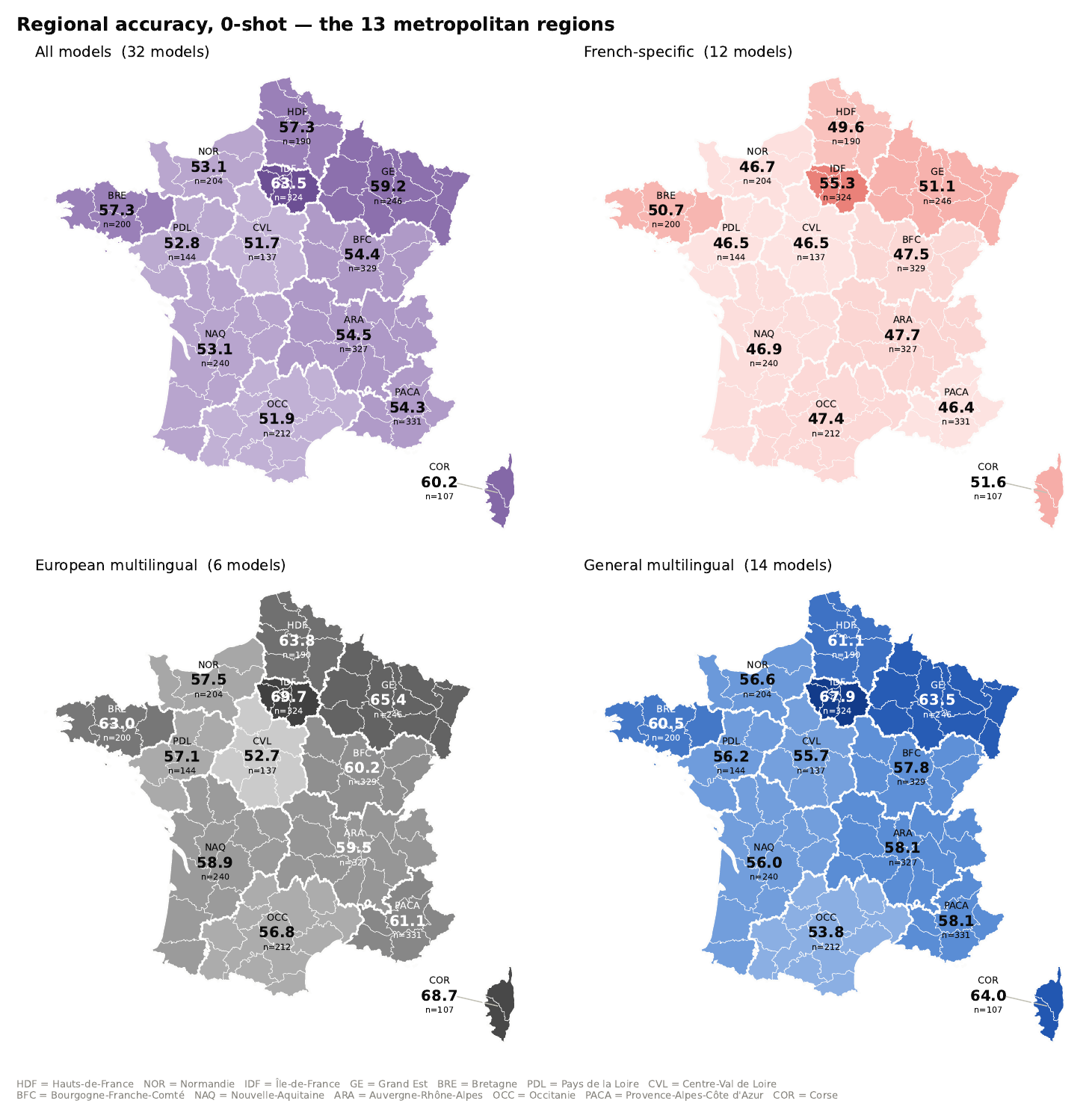}
 \caption{\new{Mean 0-shot accuracy per metropolitan region on \benchmark\ (\numberquestions\ questions), over all 32 open-weight models run locally and separately for the three model families. Department boundaries are drawn as thin lines and colored by their region; each panel uses its own single-hue scale, darker meaning more accurate.}}
  \label{fig:regional_map}
\end{figure*}

\new{Table~\ref{tab:CartePerRegionAccuracy} gives the accuracy of every local model on the questions of each region, Figure~\ref{fig:regional_map} maps the mean accuracy per region for all local models and for each family, and Figure~\ref{fig:winners} the best model per region within each family.}

\begin{table*}[t]
\centering
\resizebox{\linewidth}{!}{\begin{tabular}{ll cccccccccccccc c}
\toprule
\multicolumn{17}{c}{\textbf{\benchmark\ (3,132 questions)}} \\
\textbf{Model} & \textbf{Type} & \textbf{ARA} & \textbf{BFC} & \textbf{BRE} & \textbf{CVL} & \textbf{COR} & \textbf{FR} & \textbf{GE} & \textbf{HdF} & \textbf{NOR} & \textbf{NA} & \textbf{OCC} & \textbf{PdL} & \textbf{PACA} & \textbf{IdF} & \textbf{Overall} \\
\midrule
\multicolumn{17}{l}{\textit{French-specific LLMs}} \\[2pt]
Luciole-23B-Instruct & IT & \textbf{76.1} & \textbf{72.9} & \textbf{79.0} & \underline{65.0} & \underline{77.6} & \underline{88.7} & \underline{77.2} & \textbf{81.6} & \underline{73.0} & \textbf{74.2} & \textbf{72.2} & \textbf{72.2} & \underline{70.1} & \textbf{81.2} & \textbf{75.6} \\
Luciole-8B-Instruct & IT & 67.0 & 66.9 & 73.0 & 59.1 & \underline{77.6} & 84.4 & 71.1 & \underline{75.8} & 65.2 & 65.4 & 69.3 & 66.0 & 65.3 & \underline{80.2} & 70.1 \\
Gaperon-1125-8B-SFT & IT & 60.6 & 62.9 & 68.0 & 58.4 & 64.5 & 76.6 & 65.0 & 65.8 & 59.8 & 60.8 & 59.9 & 56.9 & 62.2 & 75.0 & 64.1 \\
Gaperon-1125-8B & Base & 57.5 & 62.6 & 61.0 & 54.7 & 57.0 & 73.8 & 65.0 & 62.1 & 62.3 & 65.0 & 57.5 & 55.6 & 62.5 & 71.9 & 62.5 \\
Claire-Mistral-7B-0.1 & Base & 45.3 & 45.6 & 51.5 & 44.5 & 54.2 & 53.2 & 51.6 & 50.0 & 47.1 & 44.6 & 42.0 & 44.4 & 45.0 & 55.9 & 48.0 \\
Lucie-7B-Instruct-v1.1 & IT & 53.2 & 50.8 & 57.0 & 42.3 & 60.7 & 63.1 & 55.7 & 57.9 & 47.1 & 53.8 & 53.8 & 45.8 & 55.9 & 62.7 & 54.5 \\
Vigogne-2-7B-Instruct & IT & 48.0 & 48.0 & 50.5 & 51.8 & 47.7 & 54.6 & 54.9 & 47.4 & 45.6 & 45.0 & 45.3 & 49.3 & 46.5 & 54.9 & 49.2 \\
Luth-LFM2-1.2B & IT & 44.3 & 45.6 & 51.0 & 51.1 & 57.9 & 61.7 & 51.6 & 44.2 & 51.5 & 42.5 & 45.3 & 48.6 & 45.0 & 54.9 & 48.8 \\
Luciole-1B-Instruct & IT & 38.5 & 41.0 & 47.5 & 47.4 & 49.5 & 53.2 & 40.7 & 41.6 & 39.2 & 40.4 & 42.9 & 33.3 & 40.8 & 45.7 & 42.4 \\
Vigogne-7B-Instruct & IT & 32.4 & 33.4 & 33.5 & 39.4 & 31.8 & 38.3 & 34.6 & 26.3 & 28.9 & 29.6 & 33.5 & 35.4 & 27.5 & 34.9 & 32.4 \\
Claire-7B-0.1 & Base & 23.9 & 18.8 & 21.5 & 21.9 & 23.4 & 18.4 & 21.1 & 22.1 & 22.5 & 24.6 & 25.9 & 26.4 & 17.5 & 24.4 & 22.1 \\
CroissantLLMChat-v0.1 & IT & 26.0 & 21.3 & 15.0 & 21.9 & 16.8 & 20.6 & 24.0 & 20.5 & 18.6 & 17.5 & 20.8 & 24.3 & 18.4 & 22.5 & 20.8 \\
\midrule
\multicolumn{17}{l}{\textit{European multilingual LLMs}} \\[2pt]
Mistral-Nemo-Instruct-2407 & IT & 64.5 & 66.0 & 73.5 & 56.2 & 73.8 & 81.6 & 67.5 & 69.5 & 59.3 & 63.8 & 67.5 & 68.8 & \textbf{71.3} & 77.5 & 68.6 \\
EuroLLM-9B-Instruct-2512 & IT & 63.0 & 62.3 & 66.5 & 53.3 & \textbf{79.4} & 83.0 & 69.9 & 69.5 & 59.8 & \underline{70.0} & 61.3 & 57.6 & 64.7 & 74.7 & 66.5 \\
Occiglot-7B-EU5-Instruct & IT & 63.3 & 61.4 & 61.5 & 51.8 & 69.2 & 78.0 & 66.3 & 64.2 & 58.3 & 58.8 & 58.0 & 54.2 & 59.2 & 67.6 & 62.2 \\
Mistral-7B-v0.1 & Base & 55.0 & 58.4 & 61.5 & 58.4 & 62.6 & 74.5 & 64.6 & 60.0 & 57.4 & 52.5 & 55.7 & 59.0 & 56.8 & 66.4 & 59.7 \\
Mistral-7B-Instruct-v0.3 & IT & 57.8 & 59.6 & 61.0 & 51.1 & 65.4 & 76.6 & 63.0 & 61.1 & 58.8 & 54.6 & 50.0 & 52.8 & 60.1 & 65.7 & 59.7 \\
Mistral-7B-Instruct-v0.2 & IT & 53.5 & 53.8 & 54.0 & 45.3 & 61.7 & 69.5 & 61.4 & 58.4 & 51.5 & 53.8 & 48.1 & 50.0 & 54.7 & 66.0 & 55.9 \\
\midrule
\multicolumn{17}{l}{\textit{General multilingual LLMs}} \\[2pt]
Qwen3.5-9B & IT & \underline{71.9} & 66.0 & 73.0 & \textbf{66.4} & 73.8 & 85.1 & 72.8 & 72.6 & \textbf{74.0} & 67.5 & \underline{70.3} & 68.1 & \textbf{71.3} & 77.2 & \underline{71.9} \\
Gemma-3-12B-IT & IT & 69.7 & \underline{70.5} & \underline{74.5} & \underline{65.0} & 71.0 & \textbf{89.4} & 74.8 & 68.9 & 63.7 & 63.3 & 67.5 & 64.6 & 67.4 & 77.5 & 70.5 \\
GLM-4.7-Flash & IT & 67.3 & \underline{70.5} & 71.5 & 63.5 & 76.6 & 84.4 & \textbf{79.7} & 67.9 & 65.7 & 66.2 & 64.6 & 65.3 & 66.8 & 79.3 & 70.6 \\
Llama-3.1-8B-Instruct & IT & 67.9 & 68.7 & 69.5 & 58.4 & 69.2 & 83.7 & 69.5 & 63.7 & 60.3 & 63.3 & 56.6 & \underline{69.4} & 68.0 & 75.3 & 67.5 \\
Qwen3.5-4B & IT & 65.7 & 62.9 & 66.5 & 62.0 & 70.1 & 78.7 & 72.4 & 72.1 & 68.1 & 63.8 & 59.0 & 63.9 & 67.1 & 72.5 & 67.3 \\
GLM-4-9B-0414 & IT & 63.6 & 64.7 & 63.0 & 56.9 & 72.9 & 76.6 & 67.5 & 68.4 & 60.3 & 61.7 & 61.3 & 65.3 & 60.4 & 74.7 & 65.3 \\
Gemma-4-E4B-IT & IT & 63.3 & 61.7 & 65.5 & 56.9 & 68.2 & 83.0 & 69.1 & 66.3 & 56.4 & 57.1 & 58.0 & 57.6 & 59.2 & 72.8 & 63.7 \\
Aya-Expanse-8B & IT & 63.6 & 61.1 & 61.5 & 56.2 & 68.2 & 74.5 & 69.5 & 66.8 & 60.8 & 59.6 & 61.8 & 56.9 & 64.0 & 70.7 & 64.0 \\
Gemma-4-E2B-IT & IT & 54.1 & 55.6 & 59.0 & 59.1 & 59.8 & 72.3 & 60.2 & 64.2 & 54.4 & 51.2 & 49.1 & 52.8 & 52.9 & 64.8 & 57.3 \\
GLM-4-9B-Chat & IT & 56.6 & 53.2 & 57.5 & 49.6 & 59.8 & 71.6 & 60.2 & 62.6 & 51.0 & 52.1 & 50.0 & 44.4 & 56.5 & 67.0 & 56.8 \\
Llama-3.2-3B-Instruct & IT & 52.3 & 49.8 & 53.0 & 49.6 & 69.2 & 61.7 & 52.8 & 53.2 & 49.5 & 47.9 & 40.1 & 50.0 & 49.2 & 65.4 & 52.7 \\
Gemma-4-E2B & Base & 37.9 & 41.9 & 45.0 & 43.8 & 37.4 & 52.5 & 50.0 & 44.7 & 39.2 & 44.2 & 41.0 & 36.8 & 44.7 & 53.1 & 44.1 \\
Llama-3.2-1B-Instruct & IT & 41.3 & 41.3 & 44.5 & 48.9 & 51.4 & 53.9 & 44.3 & 39.5 & 42.6 & 43.8 & 33.0 & 45.8 & 45.0 & 51.5 & 44.3 \\
BLOOMZ-3B & IT & 38.8 & 41.6 & 42.5 & 43.1 & 48.6 & 57.4 & 45.9 & 44.2 & 46.6 & 42.1 & 40.6 & 46.5 & 41.4 & 49.4 & 44.2 \\
\bottomrule
\end{tabular}}
\caption{Model accuracy (\%) per region on \benchmark\, 0-shot. \textbf{Bold} and \underline{underline} denote the highest and second-highest values per column. ARA: Auvergne-Rhône-Alpes, BFC: Bourgogne-Franche-Comté, BRE: Bretagne, CVL: Centre-Val de Loire, COR: Corse, FR: France (national scope), GE: Grand Est, HdF: Hauts-de-France, NOR: Normandie, NA: Nouvelle-Aquitaine, OCC: Occitanie, PdL: Pays de la Loire, PACA: Provence-Alpes-Côte d'Azur, IdF: Île-de-France.}
\label{tab:CartePerRegionAccuracy}
\end{table*}

\subsection[Regional and Topical Disparities]{\new{Regional and Topical Disparities}}

\begin{figure*}[t]
    \centering
    \includegraphics[width=\textwidth]{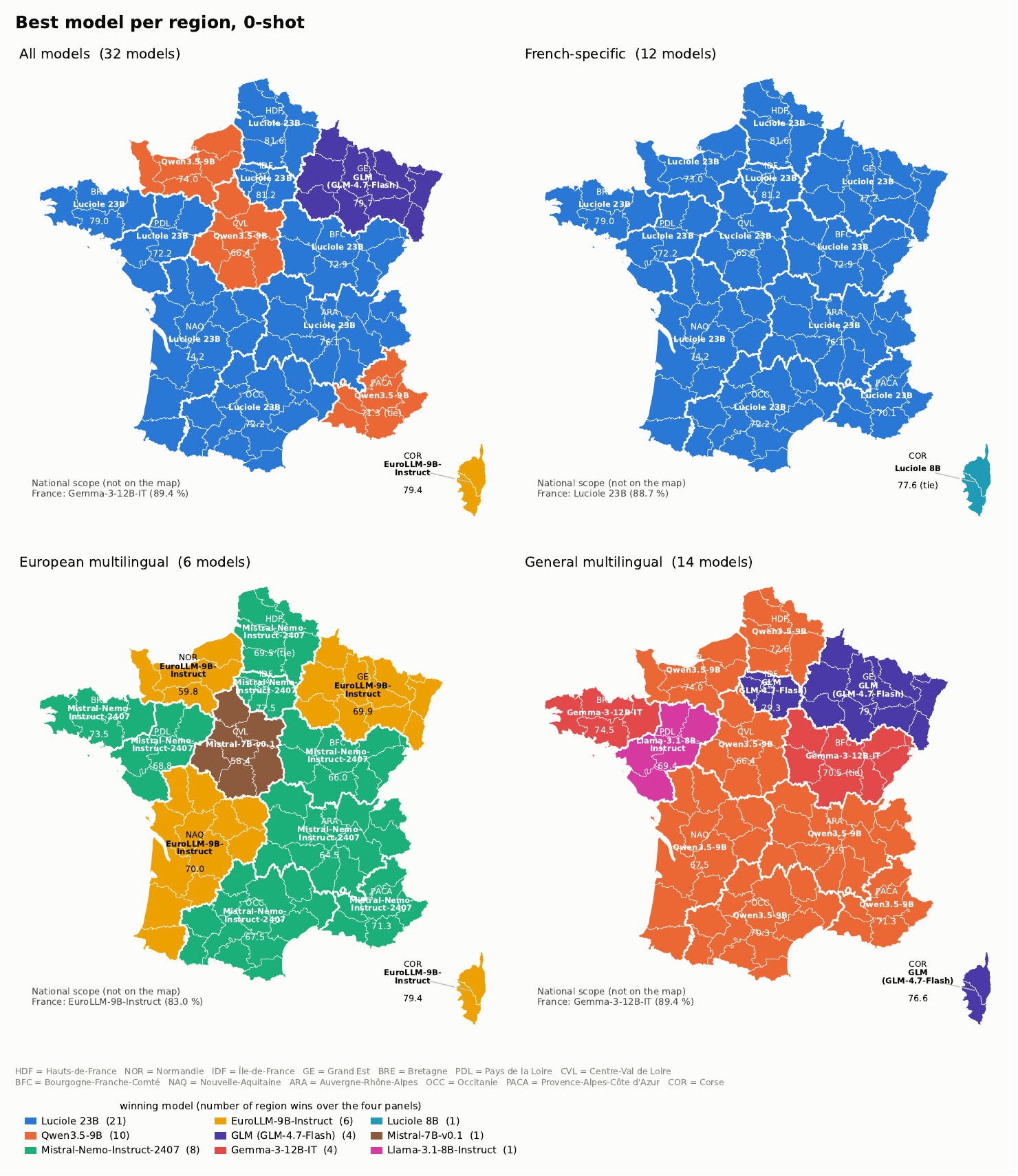}
    \caption{\new{Best model per region at 0-shot on \benchmark, for all 32 local models and within each family. Measured value is the winner's accuracy on the questions about that region.}}
    \label{fig:winners}
\end{figure*}

\begin{figure*}[t!]
  \centering
  \includegraphics[width=\textwidth]{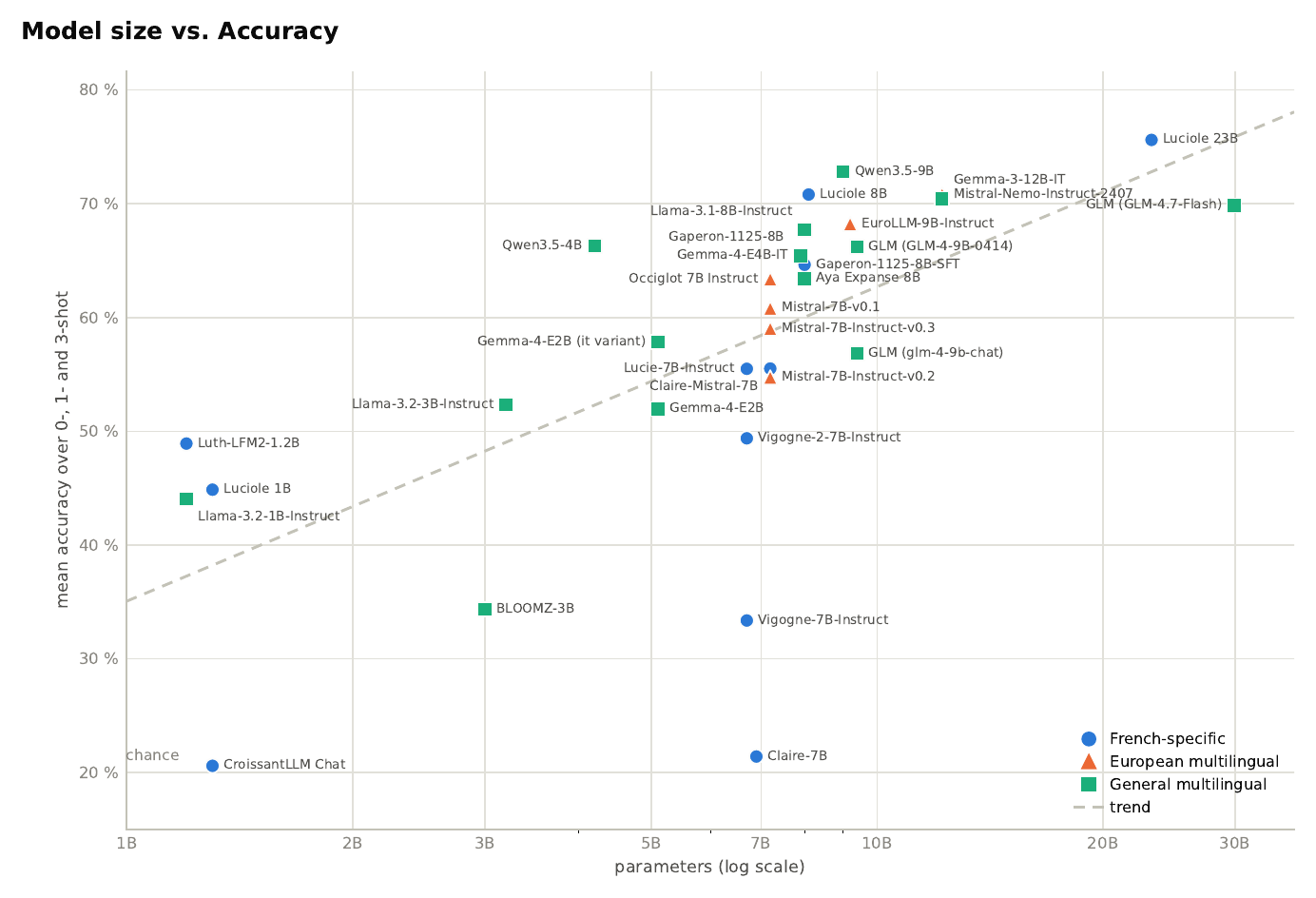}
 \caption{\new{Parameter count against mean accuracy over the three shot settings for the 32 open-weight models, with a least-squares trend in $\log_{10}$(parameters). GLM-4.7-Flash is a 30B mixture of experts with about 3B active parameters.}}
  \label{fig:acc_vs_size}
\end{figure*}

\new{Figure~\ref{fig:regional_map} and Table~\ref{tab:CartePerRegionAccuracy} show a consistent geography of model knowledge. The capital region and the national level are better covered, while the other regions are less known. The ordering is stable across the three model families, so it reflects the coverage of the regions in pretraining data rather than the behavior of a few models. Figure~\ref{fig:winners} charts the best model for each region.} Lastly, Figure~\ref{fig:acc_vs_size} plots accuracy against model size. 

\new{Per topic (Table~\ref{tab:CartePerTopicAccuracy}), institutions and governance is the best-known domain, followed by society and infrastructure; demography, territorial planning, history and heritage, and agriculture and terroirs are the least known among the well-populated topics.}

\section[Linguistic Variation Results]{\new{Linguistic Variation Results}}
\label{ap:lvresults}

\new{On the \nLV\ \Lbench\ questions (Tables~\ref{tab:LVPerRegionAccuracy} and \ref{tab:lv_model_accuracy}) the subset is slightly easier than the full benchmark for the European and general multilingual families and as hard for the French-specific one. This suggests that the French-specific models have no systematic advantage on regional lexicon over general multilingual models of the same size, which suggests that regional varieties of French are as scarce in French-centric pretraining corpora as in multilingual ones.}

\new{Table~\ref{tab:LVPerRegionAccuracy} gives the accuracy per region on the subset and Table~\ref{tab:lv_model_accuracy} the accuracy of every local model across shot settings.}
\begin{table}[h]
\centering
\resizebox{\columnwidth}{!}{%
\begin{tabular}{ll ccc ccc}
\toprule
\textbf{Model} & \textbf{Type} & \multicolumn{3}{c}{\textbf{\Lbench}} & \multicolumn{3}{c}{\textbf{\benchmark}} \\
\cmidrule(lr){3-5} \cmidrule(lr){6-8}
 & & \textbf{0-shot} & \textbf{1-shot} & \textbf{3-shot} & \textbf{0-shot} & \textbf{1-shot} & \textbf{3-shot} \\
\midrule
\multicolumn{8}{l}{\textit{French-specific LLMs}} \\[2pt]
Luciole-23B-Instruct & IT & \underline{77.3} & 75.9 & 75.5 & \textbf{75.6} & \textbf{75.3} & \textbf{75.9} \\
Luciole-8B-Instruct & IT & 74.1 & 71.8 & 76.4 & 70.1 & 70.2 & 72.2 \\
Gaperon-1125-8B-SFT & IT & 67.3 & 62.3 & 67.3 & 64.1 & 63.3 & 66.5 \\
Gaperon-1125-8B & Base & 60.9 & 64.5 & 67.3 & 62.5 & 63.7 & 67.3 \\
Claire-Mistral-7B-0.1 & Base & 36.8 & 60.9 & 59.5 & 48.0 & 58.4 & 60.1 \\
Lucie-7B-Instruct-v1.1 & IT & 48.2 & 58.2 & 60.5 & 54.5 & 55.0 & 56.9 \\
Vigogne-2-7B-Instruct & IT & 48.6 & 51.8 & 50.0 & 49.2 & 49.4 & 49.6 \\
Luth-LFM2-1.2B & IT & 55.9 & 49.5 & 51.8 & 48.8 & 48.8 & 49.2 \\
Luciole-1B-Instruct & IT & 43.2 & 48.6 & 50.0 & 42.4 & 46.0 & 46.2 \\
Vigogne-7B-Instruct & IT & 26.8 & 33.6 & 36.4 & 32.4 & 34.1 & 33.6 \\
Claire-7B-0.1 & Base & 21.8 & 26.4 & 19.1 & 22.1 & 21.3 & 20.8 \\
CroissantLLMChat-v0.1 & IT & 22.7 & 24.1 & 21.4 & 20.8 & 20.6 & 20.4 \\
\midrule
\multicolumn{8}{l}{\textit{European multilingual LLMs}} \\[2pt]
Mistral-Nemo-Instruct-2407 & IT & \textbf{78.6} & \underline{77.7} & \underline{77.3} & 68.6 & 71.5 & 72.4 \\
EuroLLM-9B-Instruct-2512 & IT & 70.0 & 66.8 & 68.6 & 66.5 & 69.2 & 69.0 \\
Occiglot-7B-EU5-Instruct & IT & 61.8 & 63.6 & 65.9 & 62.2 & 63.7 & 64.2 \\
Mistral-7B-v0.1 & Base & 59.1 & 63.2 & 63.6 & 59.7 & 61.5 & 61.2 \\
Mistral-7B-Instruct-v0.3 & IT & 59.1 & 56.8 & 56.4 & 59.7 & 59.3 & 58.0 \\
Mistral-7B-Instruct-v0.2 & IT & 60.5 & 54.1 & 55.9 & 55.9 & 53.9 & 54.4 \\
\midrule
\multicolumn{8}{l}{\textit{General multilingual LLMs}} \\[2pt]
Qwen3.5-9B & IT & 71.4 & \textbf{80.0} & \underline{77.3} & \underline{71.9} & \underline{73.0} & \underline{73.6} \\
Gemma-3-12B-IT & IT & 73.2 & 70.0 & 71.8 & 70.5 & 70.0 & 70.8 \\
GLM-4.7-Flash & IT & 70.0 & 68.6 & 70.9 & 70.6 & 69.0 & 70.0 \\
Llama-3.1-8B-Instruct & IT & 70.0 & 71.8 & 72.3 & 67.5 & 67.7 & 68.0 \\
Qwen3.5-4B & IT & 66.4 & 65.9 & 66.4 & 67.3 & 66.0 & 65.6 \\
GLM-4-9B-0414 & IT & 66.8 & 69.1 & 67.3 & 65.3 & 66.4 & 67.0 \\
Gemma-4-E4B-IT & IT & 74.1 & 76.8 & \textbf{77.7} & 63.7 & 65.6 & 67.0 \\
Aya-Expanse-8B & IT & 71.8 & 68.6 & 68.6 & 64.0 & 62.5 & 63.6 \\
Gemma-4-E2B-IT & IT & 60.9 & 64.1 & 65.9 & 57.3 & 58.2 & 58.0 \\
GLM-4-9B-Chat & IT & 59.5 & 57.3 & 57.7 & 56.8 & 56.9 & 56.9 \\
Llama-3.2-3B-Instruct & IT & 55.9 & 51.4 & 54.1 & 52.7 & 51.7 & 52.7 \\
Gemma-4-E2B & Base & 50.0 & 60.0 & 65.9 & 44.1 & 54.2 & 57.6 \\
Llama-3.2-1B-Instruct & IT & 50.5 & 49.1 & 48.6 & 44.3 & 43.8 & 44.2 \\
BLOOMZ-3B & IT & 50.9 & 33.2 & 38.2 & 44.2 & 26.3 & 32.6 \\
\bottomrule
\end{tabular}%
}
\caption{Accuracy (\%) on the \Lbench\ subset (220 questions) and on the full \benchmark\ across shot settings. \textbf{Bold} and \underline{underline} denote the highest and second-highest values per column.}
\label{tab:lv_model_accuracy}
\end{table}

\begin{table*}[t]
\centering
\resizebox{\linewidth}{!}{\begin{tabular}{ll ccccccccccccc c}
\toprule
\multicolumn{16}{c}{\textbf{\Lbench\ (220 questions)}} \\
\textbf{Model} & \textbf{Type} & \textbf{ARA} & \textbf{BFC} & \textbf{BRE} & \textbf{CVL} & \textbf{COR} & \textbf{GE} & \textbf{HdF} & \textbf{NOR} & \textbf{NA} & \textbf{OCC} & \textbf{PdL} & \textbf{PACA} & \textbf{IdF} & \textbf{Overall} \\
\midrule
\multicolumn{16}{l}{\textit{French-specific LLMs}} \\[2pt]
Luciole-23B-Instruct & IT & \underline{95.0} & \underline{95.0} & \underline{81.8} & 47.1 & \textbf{100.0} & 76.2 & \underline{85.0} & \underline{81.8} & 55.6 & \underline{76.2} & \textbf{53.3} & \underline{72.7} & 76.5 & \underline{77.3} \\
Luciole-8B-Instruct & IT & 90.0 & 85.0 & \underline{81.8} & \underline{58.8} & \textbf{100.0} & 71.4 & 70.0 & \textbf{90.9} & 55.6 & 61.9 & \textbf{53.3} & 63.6 & 82.4 & 74.1 \\
Gaperon-1125-8B-SFT & IT & 90.0 & 75.0 & \underline{81.8} & \underline{58.8} & \underline{94.4} & 52.4 & 45.0 & 72.7 & 44.4 & 66.7 & 40.0 & \underline{72.7} & \underline{88.2} & 67.3 \\
Gaperon-1125-8B & Base & 65.0 & 70.0 & 63.6 & \textbf{70.6} & 66.7 & 47.6 & 50.0 & \underline{81.8} & 55.6 & 57.1 & 26.7 & \underline{72.7} & 76.5 & 60.9 \\
Claire-Mistral-7B-0.1 & Base & 50.0 & 45.0 & 54.5 & 29.4 & 66.7 & 28.6 & 30.0 & 54.5 & 22.2 & 23.8 & 26.7 & 18.2 & 35.3 & 36.8 \\
Lucie-7B-Instruct-v1.1 & IT & 80.0 & 60.0 & \underline{81.8} & 23.5 & 66.7 & 33.3 & 45.0 & 54.5 & 16.7 & 47.6 & 26.7 & 63.6 & 41.2 & 48.2 \\
Vigogne-2-7B-Instruct & IT & 55.0 & 75.0 & 63.6 & 41.2 & 50.0 & 42.9 & 45.0 & 54.5 & 16.7 & 52.4 & \textbf{53.3} & 27.3 & 52.9 & 48.6 \\
Luth-LFM2-1.2B & IT & 75.0 & 75.0 & 45.5 & 23.5 & 77.8 & 57.1 & 45.0 & \underline{81.8} & 38.9 & 47.6 & 33.3 & 63.6 & 64.7 & 55.9 \\
Luciole-1B-Instruct & IT & 55.0 & 45.0 & 36.4 & 47.1 & 55.6 & 38.1 & 40.0 & 63.6 & 33.3 & 42.9 & 20.0 & 45.5 & 41.2 & 43.2 \\
Vigogne-7B-Instruct & IT & 20.0 & 40.0 & 9.1 & 35.3 & 33.3 & 33.3 & 35.0 & 27.3 & 5.6 & 23.8 & 26.7 & 27.3 & 23.5 & 26.8 \\
Claire-7B-0.1 & Base & 15.0 & 20.0 & 18.2 & 23.5 & 27.8 & 28.6 & 30.0 & 27.3 & 27.8 & 9.5 & 20.0 & 18.2 & 17.6 & 21.8 \\
CroissantLLMChat-v0.1 & IT & 35.0 & 35.0 & 9.1 & 23.5 & 16.7 & 19.0 & 15.0 & 27.3 & 22.2 & 28.6 & 20.0 & 0.0 & 29.4 & 22.7 \\
\midrule
\multicolumn{16}{l}{\textit{European multilingual LLMs}} \\[2pt]
Mistral-Nemo-Instruct-2407 & IT & \underline{95.0} & \textbf{100.0} & \underline{81.8} & 47.1 & \underline{94.4} & 71.4 & 80.0 & 72.7 & \textbf{72.2} & \textbf{81.0} & \underline{46.7} & \underline{72.7} & \textbf{94.1} & \textbf{78.6} \\
EuroLLM-9B-Instruct-2512 & IT & \underline{95.0} & 90.0 & 72.7 & 47.1 & \underline{94.4} & 66.7 & 75.0 & \textbf{90.9} & \underline{66.7} & 57.1 & 26.7 & 45.5 & 70.6 & 70.0 \\
Occiglot-7B-EU5-Instruct & IT & 75.0 & \textbf{100.0} & 63.6 & 23.5 & 77.8 & 66.7 & 70.0 & 63.6 & 61.1 & 52.4 & 20.0 & 54.5 & 58.8 & 61.8 \\
Mistral-7B-v0.1 & Base & 65.0 & \underline{95.0} & 63.6 & 47.1 & 72.2 & 57.1 & 65.0 & 63.6 & 27.8 & 52.4 & 40.0 & 36.4 & 70.6 & 59.1 \\
Mistral-7B-Instruct-v0.3 & IT & 90.0 & \underline{95.0} & 72.7 & 23.5 & 72.2 & 61.9 & 45.0 & 63.6 & 38.9 & 47.6 & 40.0 & 36.4 & 70.6 & 59.1 \\
Mistral-7B-Instruct-v0.2 & IT & \underline{95.0} & \underline{95.0} & 63.6 & 41.2 & 72.2 & 76.2 & 45.0 & 54.5 & 33.3 & 52.4 & 40.0 & 36.4 & 58.8 & 60.5 \\
\midrule
\multicolumn{16}{l}{\textit{General multilingual LLMs}} \\[2pt]
Qwen3.5-9B & IT & \underline{95.0} & \underline{95.0} & \underline{81.8} & 41.2 & \underline{94.4} & 81.0 & 80.0 & \underline{81.8} & 44.4 & 52.4 & 40.0 & 54.5 & 76.5 & 71.4 \\
Gemma-3-12B-IT & IT & \underline{95.0} & \textbf{100.0} & \textbf{90.9} & 29.4 & 88.9 & 81.0 & 65.0 & \underline{81.8} & \textbf{72.2} & 66.7 & 33.3 & 63.6 & 76.5 & 73.2 \\
GLM-4.7-Flash & IT & \underline{95.0} & \textbf{100.0} & \underline{81.8} & 35.3 & \underline{94.4} & \underline{85.7} & 55.0 & \textbf{90.9} & 38.9 & 57.1 & 33.3 & 63.6 & 76.5 & 70.0 \\
Llama-3.1-8B-Instruct & IT & \underline{95.0} & \textbf{100.0} & \underline{81.8} & 41.2 & 83.3 & 71.4 & 65.0 & 63.6 & 55.6 & 57.1 & \underline{46.7} & 63.6 & 76.5 & 70.0 \\
Qwen3.5-4B & IT & \underline{95.0} & \underline{95.0} & \underline{81.8} & 35.3 & 88.9 & 71.4 & 60.0 & \textbf{90.9} & 27.8 & 38.1 & 33.3 & \underline{72.7} & 82.4 & 66.4 \\
GLM-4-9B-0414 & IT & 85.0 & 85.0 & 72.7 & 29.4 & 83.3 & 61.9 & 60.0 & \textbf{90.9} & 33.3 & 61.9 & \textbf{53.3} & \underline{72.7} & \underline{88.2} & 66.8 \\
Gemma-4-E4B-IT & IT & \textbf{100.0} & \underline{95.0} & \underline{81.8} & 41.2 & 88.9 & \textbf{90.5} & \textbf{90.0} & \textbf{90.9} & 50.0 & 52.4 & 26.7 & \textbf{81.8} & 70.6 & 74.1 \\
Aya-Expanse-8B & IT & 90.0 & \textbf{100.0} & \underline{81.8} & 47.1 & 83.3 & 76.2 & 70.0 & \underline{81.8} & \underline{66.7} & 61.9 & 26.7 & 54.5 & 82.4 & 71.8 \\
Gemma-4-E2B-IT & IT & 70.0 & 90.0 & 72.7 & 35.3 & 72.2 & 42.9 & 75.0 & \underline{81.8} & 61.1 & 47.6 & 33.3 & 54.5 & 58.8 & 60.9 \\
GLM-4-9B-Chat & IT & 85.0 & 85.0 & 63.6 & 29.4 & 66.7 & 47.6 & 60.0 & \textbf{90.9} & 38.9 & 47.6 & 40.0 & 63.6 & 64.7 & 59.5 \\
Llama-3.2-3B-Instruct & IT & 85.0 & 85.0 & 72.7 & 41.2 & 77.8 & 52.4 & 55.0 & \underline{81.8} & 22.2 & 23.8 & 33.3 & 18.2 & 76.5 & 55.9 \\
Gemma-4-E2B & Base & 60.0 & 70.0 & \underline{81.8} & 35.3 & 55.6 & 47.6 & 40.0 & 54.5 & 44.4 & 38.1 & 26.7 & 54.5 & 52.9 & 50.0 \\
Llama-3.2-1B-Instruct & IT & 65.0 & 65.0 & 45.5 & 47.1 & 66.7 & 61.9 & 45.0 & 63.6 & 22.2 & 28.6 & 40.0 & 54.5 & 52.9 & 50.5 \\
BLOOMZ-3B & IT & 55.0 & 80.0 & 63.6 & 41.2 & 44.4 & 52.4 & 50.0 & 63.6 & 22.2 & 33.3 & \underline{46.7} & 63.6 & 58.8 & 50.9 \\
\bottomrule
\end{tabular}}
\caption{Model accuracy (\%) per region on the \Lbench\ subset, 0-shot. \textbf{Bold} and \underline{underline} denote the highest and second-highest values per column. ARA: Auvergne-Rhône-Alpes, BFC: Bourgogne-Franche-Comté, BRE: Bretagne, CVL: Centre-Val de Loire, COR: Corse, FR: France (national scope), GE: Grand Est, HdF: Hauts-de-France, NOR: Normandie, NA: Nouvelle-Aquitaine, OCC: Occitanie, PdL: Pays de la Loire, PACA: Provence-Alpes-Côte d'Azur, IdF: Île-de-France.}
\label{tab:LVPerRegionAccuracy}
\end{table*}

\begin{table*}[h]
\centering
\resizebox{\textwidth}{!}{%
\begin{tabular}{ll cccccccccccc c}
\toprule
\textbf{Model} & \textbf{Type} & \textbf{Agri} & \textbf{Culture} & \textbf{Démo} & \textbf{Enviro} & \textbf{Hist} & \textbf{Infra} & \textbf{Inst} & \textbf{Lang} & \textbf{Soc} & \textbf{Terr} & \textbf{Transp} & \textbf{Éco} & \textbf{Overall} \\
\midrule
\multicolumn{15}{l}{\textit{French-specific LLMs}} \\[2pt]
Luciole-23B-Instruct & IT & \textbf{76.3} & \textbf{79.6} & 64.2 & \textbf{77.9} & \textbf{70.5} & \textbf{77.2} & \textbf{86.4} & \underline{78.7} & \textbf{83.2} & \textbf{77.6} & 67.2 & \underline{70.4} & \textbf{75.6} \\
Luciole-8B-Instruct & IT & \underline{72.3} & \underline{77.0} & 53.9 & 68.3 & 69.0 & 72.5 & 80.4 & 75.5 & 75.5 & \underline{72.9} & 56.9 & 58.4 & 70.1 \\
Gaperon-1125-8B-SFT & IT & 57.2 & 66.7 & 55.9 & 62.4 & 61.8 & 70.1 & 78.5 & 67.1 & 69.0 & 63.6 & 63.8 & 58.1 & 64.1 \\
Gaperon-1125-8B & Base & 62.4 & 64.3 & 58.8 & 67.2 & 57.0 & 71.3 & 75.2 & 61.7 & 68.4 & 55.1 & 65.5 & 56.1 & 62.5 \\
Claire-Mistral-7B-0.1 & Base & 48.6 & 46.3 & 43.6 & 51.3 & 45.2 & 50.9 & 61.7 & 40.4 & 49.7 & 47.7 & 58.6 & 48.9 & 48.0 \\
Lucie-7B-Instruct-v1.1 & IT & 54.9 & 63.1 & 36.3 & 56.5 & 52.8 & 62.3 & 72.4 & 48.7 & 58.1 & 50.5 & 43.1 & 44.7 & 54.5 \\
Vigogne-2-7B-Instruct & IT & 41.0 & 51.5 & 48.5 & 50.6 & 44.0 & 49.1 & 63.6 & 48.7 & 52.3 & 36.4 & 37.9 & 53.6 & 49.2 \\
Luth-LFM2-1.2B & IT & 39.3 & 43.8 & 48.5 & 51.7 & 39.9 & 53.9 & 59.8 & 56.7 & 63.2 & 47.7 & 53.4 & 52.0 & 48.8 \\
Luciole-1B-Instruct & IT & 37.0 & 43.0 & 41.2 & 42.1 & 36.9 & 53.3 & 49.1 & 44.0 & 47.7 & 40.2 & 32.8 & 43.6 & 42.4 \\
Vigogne-7B-Instruct & IT & 30.6 & 30.7 & 26.0 & 36.2 & 29.8 & 36.5 & 42.1 & 27.8 & 40.0 & 26.2 & 31.0 & 36.0 & 32.4 \\
Claire-7B-0.1 & Base & 21.4 & 21.9 & 22.5 & 19.9 & 22.5 & 19.8 & 21.5 & 20.6 & 22.6 & 25.2 & 19.0 & 25.7 & 22.1 \\
CroissantLLMChat-v0.1 & IT & 23.1 & 19.3 & 18.6 & 17.7 & 21.5 & 23.4 & 21.0 & 20.6 & 21.9 & 17.8 & 24.1 & 23.5 & 20.8 \\
\midrule
\multicolumn{15}{l}{\textit{European multilingual LLMs}} \\[2pt]
Mistral-Nemo-Instruct-2407 & IT & 66.5 & 72.2 & 57.4 & 69.0 & 66.6 & 69.5 & 78.0 & \textbf{79.4} & 76.8 & 66.4 & 69.0 & 55.9 & 68.6 \\
EuroLLM-9B-Instruct-2512 & IT & 69.4 & 74.6 & 49.0 & 64.9 & 67.2 & 65.3 & 79.4 & 72.6 & 68.4 & 65.4 & 55.2 & 52.0 & 66.5 \\
Occiglot-7B-EU5-Instruct & IT & 54.3 & 65.3 & 52.9 & 64.6 & 57.5 & 64.7 & 75.7 & 63.2 & 65.8 & 55.1 & 60.3 & 63.7 & 62.2 \\
Mistral-7B-v0.1 & Base & 54.3 & 60.5 & 51.5 & 60.1 & 54.5 & 62.3 & 72.4 & 61.4 & 69.7 & 51.4 & 63.8 & 61.5 & 59.7 \\
Mistral-7B-Instruct-v0.3 & IT & 51.4 & 61.8 & 53.4 & 57.6 & 57.6 & 62.3 & 75.2 & 61.7 & 65.2 & 55.1 & 58.6 & 56.7 & 59.7 \\
Mistral-7B-Instruct-v0.2 & IT & 45.1 & 59.4 & 46.1 & 54.2 & 56.8 & 57.5 & 70.6 & 60.3 & 54.8 & 47.7 & 56.9 & 51.1 & 55.9 \\
\midrule
\multicolumn{15}{l}{\textit{General multilingual LLMs}} \\[2pt]
Qwen3.5-9B & IT & 68.8 & 74.6 & \textbf{68.6} & \underline{72.0} & 66.4 & 75.4 & \underline{83.6} & 73.3 & 74.8 & 63.6 & \underline{70.7} & \textbf{71.8} & \underline{71.9} \\
Gemma-3-12B-IT & IT & 64.2 & 74.1 & \textbf{68.6} & 68.6 & 66.7 & 73.7 & 82.7 & 73.3 & 76.1 & 57.0 & \underline{70.7} & 67.3 & 70.5 \\
GLM-4.7-Flash & IT & 63.0 & 75.4 & 60.8 & 71.2 & \underline{69.5} & \underline{76.0} & \underline{83.6} & 71.5 & 72.9 & 64.5 & 63.8 & 64.5 & 70.6 \\
Llama-3.1-8B-Instruct & IT & 54.9 & 69.9 & 64.2 & 67.9 & 63.6 & 66.5 & 78.5 & 70.4 & \underline{77.4} & 66.4 & 67.2 & 66.2 & 67.5 \\
Qwen3.5-4B & IT & 64.2 & 68.2 & \underline{64.7} & 71.6 & 59.4 & 69.5 & \underline{83.6} & 69.7 & 72.9 & 56.1 & \textbf{72.4} & 66.2 & 67.3 \\
GLM-4-9B-0414 & IT & 60.7 & 66.4 & 60.8 & 69.7 & 59.8 & 67.1 & 79.9 & 68.2 & 74.8 & 52.3 & 65.5 & 62.0 & 65.3 \\
Gemma-4-E4B-IT & IT & 60.7 & 69.7 & 55.9 & 59.8 & 56.6 & 67.1 & 76.2 & 73.6 & 71.0 & 55.1 & 58.6 & 58.9 & 63.7 \\
Aya-Expanse-8B & IT & 56.1 & 67.1 & 55.4 & 64.6 & 57.0 & 70.1 & 77.1 & 70.0 & 71.0 & 53.3 & 60.3 & 65.4 & 64.0 \\
Gemma-4-E2B-IT & IT & 53.8 & 59.9 & 54.4 & 59.0 & 45.9 & 64.7 & 66.8 & 63.2 & 63.2 & 57.0 & 58.6 & 58.1 & 57.3 \\
GLM-4-9B-Chat & IT & 53.2 & 63.1 & 50.0 & 57.6 & 53.8 & 58.1 & 73.4 & 60.3 & 61.3 & 50.5 & 41.4 & 46.4 & 56.8 \\
Llama-3.2-3B-Instruct & IT & 42.8 & 55.9 & 52.0 & 55.4 & 48.3 & 56.3 & 65.4 & 55.2 & 61.9 & 48.6 & 37.9 & 46.4 & 52.7 \\
Gemma-4-E2B & Base & 41.0 & 46.5 & 40.7 & 40.6 & 37.4 & 49.1 & 54.7 & 49.8 & 47.1 & 35.5 & 53.4 & 44.1 & 44.1 \\
Llama-3.2-1B-Instruct & IT & 34.1 & 46.0 & 44.1 & 50.2 & 34.9 & 43.1 & 52.8 & 52.3 & 51.0 & 40.2 & 39.7 & 46.1 & 44.3 \\
BLOOMZ-3B & IT & 38.2 & 40.8 & 42.6 & 51.3 & 32.0 & 55.1 & 61.2 & 50.2 & 54.8 & 40.2 & 43.1 & 45.3 & 44.2 \\
\midrule
\textit{\# questions} &  & 173 & 544 & 204 & 271 & 604 & 167 & 214 & 277 & 155 & 107 & 58 & 358 & 3132 \\
\bottomrule
\end{tabular}%
}
\caption{\benchmark\ model accuracy (\%) per topic, 0-shot. Agri: Agriculture \& terroirs, Culture: Culture, traditions \& société, Démo: Démographie, Enviro: Environnement \& biodiversité, Hist: Histoire \& patrimoine, Infra: Infrastructure \& réseaux, Inst: Institutions \& gouvernance, Lang: Langue \& sciences du langage, Soc: Société \& réalités sociales, Terr: Territoire \& aménagement, Transp: Transport \& mobilité, Éco: Économie \& industrie. \textbf{Bold} and \underline{underline} denote the highest and second-highest values per column; the last row gives the number of questions per topic.}
\label{tab:CartePerTopicAccuracy}
\end{table*}


\section[Difficulty per Region]{\new{Difficulty per Region}}
\label{ap:difficulty}

\begin{figure}[ht]
    \centering
    \includegraphics[width=\columnwidth]{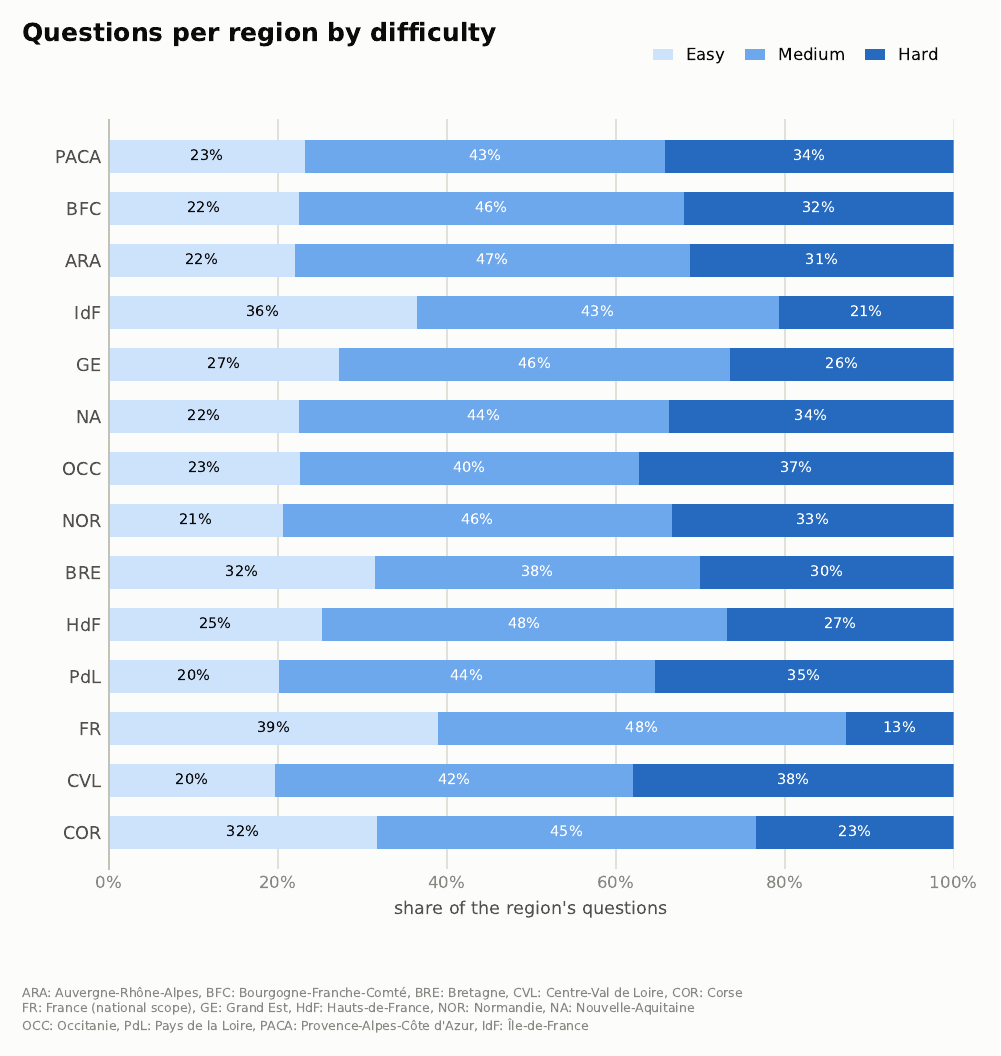}
    \caption{\new{Difficulty tier shares of \benchmark\ per region.}}
    \label{fig:difficulty_by_region}
\end{figure}

\new{Figure~\ref{fig:difficulty_by_region} gives the number and share of Easy, Medium and Hard questions in each region after the second review. Additionally, examples of questions from each difficulty tier are shown in Table~\ref{tab:question_examples_diff}.}

   \begin{table*}[ht!]
\centering
\begin{tabular}{l p{0.45\textwidth} p{0.35\textwidth}}
\toprule
\textbf{Difficulty} & \textbf{Question} & \textbf{Options} \\
\midrule
Easy & Quelle forêt d'Ille-et-Vilaine est traditionnellement associée aux légendes arthuriennes et au tombeau de Merlin ? (Which forest in Ille-et-Vilaine is traditionally associated with Arthurian legends and Merlin’s tomb?) & \textbf{A. La forêt de Brocéliande} \newline B. La forêt des Landes \newline C. Je ne sais pas \newline D. La forêt de Compiègne \newline E. La forêt de Tronçais \\
\midrule
Medium & En 2011, quel secteur industriel était particulièrement surreprésenté au sein des Entreprises de Taille Intermédiaire (ETI) de la zone d'Oyonnax ? (In 2011, which industrial sector was particularly overrepresented among the Intermediate-Sized Enterprises (ETIs) in the Oyonnax area?) & A. La construction navale \newline B. Je ne sais pas \newline \textbf{C. La plasturgie} \newline D. L'horlogerie \newline E. L'agroalimentaire \\
\midrule
Hard & Quel territoire annexé en 1768 a conservé l'usage de l'italien dans ses actes notariés jusqu'au milieu du XIXe siècle malgré la pression française ? (Which territory annexed in 1768 retained the use of Italian in its notarial records until the mid-19th century despite French pressure?) & A. La Savoie \newline B. Je ne sais pas \newline C. Le Comté de Nice \newline D. La Lorraine \newline \textbf{E. La Corse} \\
\bottomrule
\end{tabular}
\caption{Examples of questions from each difficulty tier. The correct option is highlighted in bold.}
\label{tab:question_examples_diff}
\end{table*}

\section{Quantifying Positional Bias}
\label{sec:bias_methodology}

\begin{figure}[h]
    \centering
    \includegraphics[width=\columnwidth]{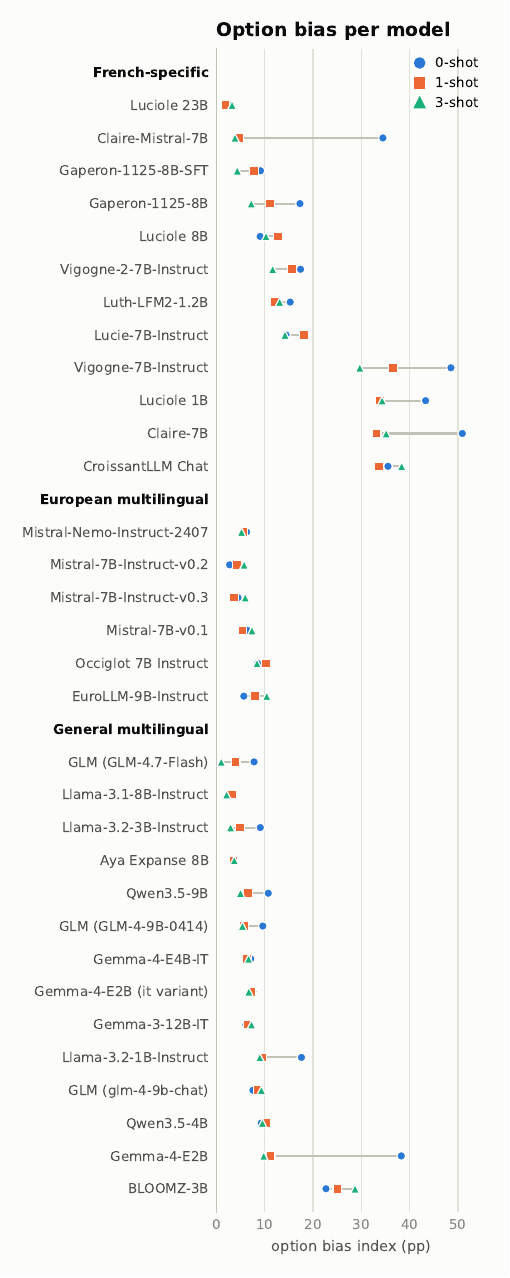}
    \caption{\new{Option bias index per model at 0, 1 and 3 shots on \benchmark\ : the share of answers (percentage points) that would have to move to another letter for the model to pick each letter exactly as often as it is the correct answer. Labels give the most over-picked letter at 0-shot and its excess over the gold share when it exceeds five points.}}
    \label{fig:llm_positional_bias}
\end{figure}

\begin{figure}[h]
    \centering
    \includegraphics[width=\columnwidth]{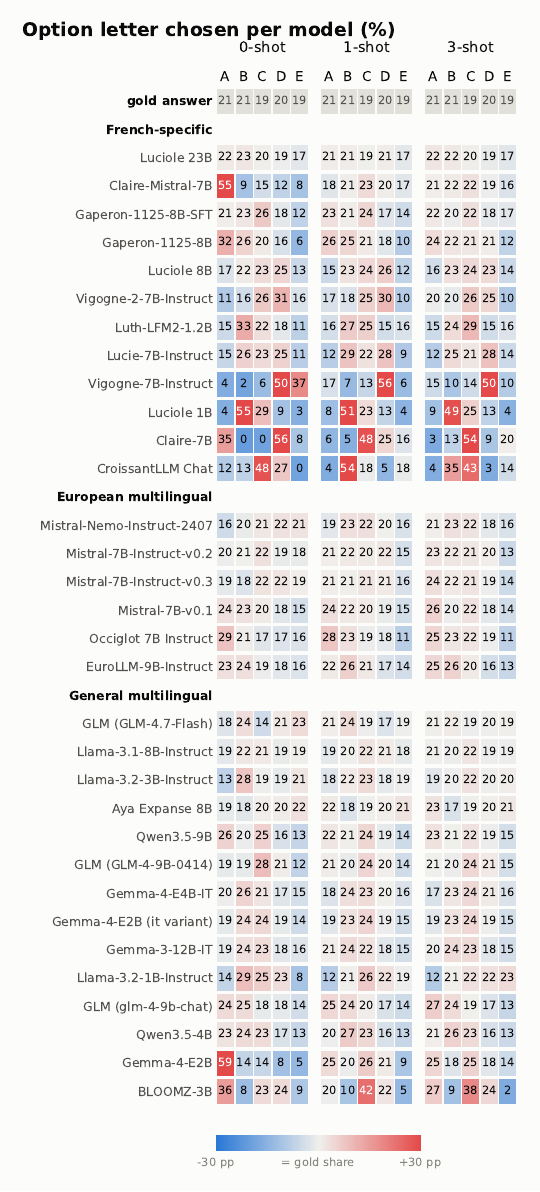}
    \caption{\new{One panel per shot setting, five columns A to E. Each cell gives the share of the 3,132 questions on which
    that letter is the model's top choice. Red means the letter is picked more often than it is the correct answer, blue less often, and the top row
    shows the gold shares for reference.}}
    \label{fig:llm_positional_bias_heatmap}
\end{figure}

To rigorously evaluate whether a given LLM exhibits a preference for specific multiple-choice positions (e.g., options A, B, C, D, or E), we quantify positional bias by comparing the distribution of a model's predicted answer letters with the distribution of correct answer letters. For each question, the predicted letter is the option with the highest log-probability; when multiple options tie for the maximum, their contribution is divided equally. Let \(p_\ell\) and \(q_\ell\) denote the proportions of predicted and correct (gold) answers, respectively, assigned to letter \(\ell \in \{A,B,C,D,E\}\). We define the option bias index as their total variation distance:
\[
\mathrm{Bias} = \frac{1}{2}\sum_{\ell=A}^{E}|p_\ell-q_\ell|.
\]
The index measures the proportion of answers that would need to be reassigned to match the gold answer-letter distribution, expressed in percentage points. A value of zero indicates matching distributions, while larger values indicate greater discrepancies in how frequently answer letters are selected. This metric captures distributional bias rather than answer accuracy: a model may have a balanced answer-letter distribution while still answering incorrectly. We report the index at 0-, 1-, and 3-shot settings in Figure~\ref{fig:llm_positional_bias}, and show per-letter deviations from the gold distribution in Figure~\ref{fig:llm_positional_bias_heatmap}.






\section[Answering Behavior]{\new{Answering Behavior}}
\label{ap:behaviour}

Table~\ref{tab:behaviour} lists the accuracy and ``I don't know'' (JNSP) rates for every local model across the 0-, 1-, and 3-shot settings, summarizing model behavior beyond binary correctness.

\paragraph{Abstention:}
Most competitive models rarely pick \textit{``Je ne sais pas''}.                                                                                                                   
\paragraph{Option Bias:}
The option bias index (Figure~\ref{fig:llm_positional_bias}; per-letter shares in Figure~\ref{fig:llm_positional_bias_heatmap}) is largest for the chance-level and base models: at 0-shot, Claire-7B mostly picks `D', CroissantLLM Chat `C', Luciole-1B `B' and the base Gemma-4-E2B `A'. Demonstrations remove most of the bias of the base models (Claire-Mistral-7B from 34.5 to 4.8 points, Gemma-4-E2B from 38.3 to 11.2) but not of the chance-level ones.

\begin{table*}[ht]
\centering
\resizebox{\textwidth}{!}{%
\begin{tabular}{ll ccc ccc ccc}
\toprule
\textbf{Model} & \textbf{Type} & \multicolumn{3}{c}{\textbf{Accuracy} $\uparrow$} & \multicolumn{3}{c}{\textbf{JNSP rate} $\downarrow$} & \multicolumn{3}{c}{\textbf{Option bias} $\downarrow$} \\
\cmidrule(lr){3-5} \cmidrule(lr){6-8} \cmidrule(lr){9-11}
 & & \textbf{0} & \textbf{1} & \textbf{3} & \textbf{0} & \textbf{1} & \textbf{3} & \textbf{0} & \textbf{1} & \textbf{3} \\
\midrule
\multicolumn{11}{l}{\textit{French-specific LLMs}} \\[2pt]
Luciole-23B-Instruct & IT & \textbf{75.6} & \textbf{75.3} & \textbf{75.9} & 3.1 & 2.4 & 2.3 & \textbf{2.7} & \textbf{1.9} & 3.4 \\
Luciole-8B-Instruct & IT & 70.1 & 70.2 & 72.2 & 11.4 & 4.5 & 3.4 & 9.0 & 12.8 & 10.1 \\
Gaperon-1125-8B-SFT & IT & 64.1 & 63.3 & 66.5 & 4.3 & 6.6 & 4.7 & 9.1 & 8.4 & 4.5 \\
Gaperon-1125-8B & Base & 62.5 & 63.7 & 67.3 & 2.4 & 5.1 & 2.2 & 17.3 & 15.9 & 11.4 \\
Claire-Mistral-7B-0.1 & Base & 48.0 & 58.4 & 60.1 & 3.3 & 0.4 & 1.1 & 34.5 & 4.8 & 4.6 \\
Lucie-7B-Instruct-v1.1 & IT & 54.5 & 55.0 & 56.9 & 18.0 & 8.0 & 7.6 & 14.4 & 18.0 & 14.5 \\
Vigogne-2-7B-Instruct & IT & 49.2 & 49.4 & 49.6 & 2.7 & 2.5 & 2.0 & 17.4 & 15.5 & 11.4 \\
Luth-LFM2-1.2B & IT & 48.8 & 48.8 & 49.2 & 7.6 & 5.2 & 8.0 & 15.3 & 12.1 & 13.5 \\
Luciole-1B-Instruct & IT & 42.4 & 46.0 & 46.2 & 7.3 & 5.5 & 5.1 & 43.3 & 34.1 & 34.5 \\
Vigogne-7B-Instruct & IT & 32.4 & 34.1 & 33.6 & 3.7 & 6.5 & 6.1 & 48.5 & 32.9 & 26.9 \\
Claire-7B-0.1 & Base & 22.1 & 21.3 & 20.8 & 14.9 & 20.8 & 21.0 & 50.9 & 33.1 & 34.5 \\
CroissantLLMChat-v0.1 & IT & 20.8 & 20.6 & 20.4 & 16.3 & 20.9 & 20.8 & 35.5 & 34.7 & 38.7 \\
\midrule
\multicolumn{11}{l}{\textit{European multilingual LLMs}} \\[2pt]
Mistral-Nemo-Instruct-2407 & IT & 68.6 & 71.5 & 72.4 & 4.5 & 0.6 & 0.8 & 6.2 & 5.9 & 5.7 \\
EuroLLM-9B-Instruct-2512 & IT & 66.5 & 69.2 & 69.0 & 12.0 & 5.7 & 7.3 & 5.7 & 8.9 & 11.1 \\
Occiglot-7B-EU5-Instruct & IT & 62.2 & 63.7 & 64.2 & 2.7 & 1.5 & 3.1 & 8.5 & 10.8 & 8.8 \\
Mistral-7B-v0.1 & Base & 59.7 & 61.5 & 61.2 & \textbf{0.1} & 1.0 & 2.4 & 6.4 & 5.4 & 7.6 \\
Mistral-7B-Instruct-v0.3 & IT & 59.7 & 59.3 & 58.0 & 3.2 & 5.1 & 8.5 & 4.4 & 3.7 & 6.0 \\
Mistral-7B-Instruct-v0.2 & IT & 55.9 & 53.9 & 54.4 & 7.4 & 10.5 & 10.6 & \textbf{2.7} & 4.2 & 5.8 \\
\midrule
\multicolumn{11}{l}{\textit{General multilingual LLMs}} \\[2pt]
Qwen3.5-9B & IT & \underline{71.9} & \underline{73.0} & \underline{73.6} & 1.0 & 0.8 & 1.8 & 10.7 & 7.2 & 5.6 \\
Gemma-3-12B-IT & IT & 70.5 & 70.0 & 70.8 & \underline{0.2} & \underline{0.3} & \textbf{0.4} & 6.1 & 6.6 & 7.3 \\
GLM-4.7-Flash & IT & 70.6 & 69.0 & 70.0 & 1.7 & 2.0 & 2.3 & 6.7 & 6.9 & 4.0 \\
Llama-3.1-8B-Instruct & IT & 67.5 & 67.7 & 68.0 & 0.5 & 0.8 & 1.4 & \underline{2.8} & \underline{2.7} & \underline{2.7} \\
Qwen3.5-4B & IT & 67.3 & 66.0 & 65.6 & 1.6 & 0.4 & 0.9 & 9.3 & 10.4 & 10.2 \\
GLM-4-9B-0414 & IT & 65.3 & 66.4 & 67.0 & 2.5 & 1.7 & 1.3 & 9.6 & 5.8 & 5.3 \\
Gemma-4-E4B-IT & IT & 63.7 & 65.6 & 67.0 & 10.5 & 7.8 & 6.3 & 7.1 & 6.6 & 6.4 \\
Aya-Expanse-8B & IT & 64.0 & 62.5 & 63.6 & 0.8 & 1.4 & 1.4 & 3.7 & 3.5 & 3.6 \\
Gemma-4-E2B-IT & IT & 57.3 & 58.2 & 58.0 & 5.0 & 2.6 & 3.7 & 7.5 & 7.2 & 6.6 \\
GLM-4-9B-Chat & IT & 56.8 & 56.9 & 56.9 & 10.0 & 9.2 & 8.6 & 7.5 & 9.1 & 10.2 \\
Llama-3.2-3B-Instruct & IT & 52.7 & 51.7 & 52.7 & 11.1 & 15.5 & 14.6 & 9.1 & 5.0 & \textbf{2.4} \\
Gemma-4-E2B & Base & 44.1 & 54.2 & 57.6 & 1.8 & \textbf{0.2} & \underline{0.7} & 38.3 & 11.2 & 10.3 \\
Llama-3.2-1B-Instruct & IT & 44.3 & 43.8 & 44.2 & 1.1 & 0.4 & \underline{0.7} & 17.6 & 9.9 & 8.2 \\
BLOOMZ-3B & IT & 44.2 & 26.3 & 32.6 & 3.1 & 26.5 & 9.9 & 22.7 & 24.9 & 28.6 \\
\bottomrule
\end{tabular}%
}
\caption{\new{Answering behaviour on \benchmark\ at 0, 1 and 3 shots. Accuracy (\%); JNSP rate: share of questions where the most probable option is \textit{``Je ne sais pas''} (always counted as wrong); option bias: total-variation distance (percentage points) between the distribution of the chosen letters and that of the gold letters, i.e.\ the share of answers that would have to move for the model to pick each letter as often as it is correct. \textbf{Bold} and \underline{underline} denote the best and second-best values per column.}}
\label{tab:behaviour}
\end{table*}







\subsection{Limitations of General-Purpose Pre-training}

Although state-of-the-art, primarily English-centric models achieve strong performance on general translated benchmarks~\cite{li2025language}, our results show a drop in accuracy when evaluation targets intra-national variation, local agriculture, regional heritage, and culturally grounded knowledge. This highlights the limitations of relying on homogenized pre-training corpora for fine-grained regional and cultural understanding. While models perform well on general French evaluations, \benchmark\ enables finer-grained analysis through topic- and region-specific breakdowns, revealing performance disparities and uneven coverage of localized knowledge. We also observe a secondary limitation in a subset of models: a tendency to favor specific answer options, suggesting that some systems may not fully rely on question-conditioned reasoning and instead exhibit label or positional biases. We further analyze this behavior in Appendix~\ref{sec:bias_methodology}.

\end{document}